\documentclass[journal]{IEEEtran}

\usepackage{times}
\usepackage{epsfig}
\usepackage{graphicx}
\usepackage{caption}
\usepackage{amsmath}
\usepackage{amssymb}
\usepackage{enumerate}

\usepackage{amsfonts}
\usepackage{multirow}
\usepackage{subfigure}
\usepackage{color}
\graphicspath{{./figures/}}
\usepackage{epstopdf}
\usepackage{float}

\usepackage[linesnumbered,ruled]{algorithm2e}

\usepackage{lineno,xcolor}
\usepackage{array}

\newcommand{\PreserveBackslash}[1]{\let\temp=\\#1\let\\=\temp}
\newcolumntype{C}[1]{>{\PreserveBackslash\centering}p{#1}}
\newcolumntype{R}[1]{>{\PreserveBackslash\raggedleft}p{#1}}
\newcolumntype{L}[1]{>{\PreserveBackslash\raggedright}p{#1}}

\usepackage{multirow}

\usepackage{amsmath,bm}  
\usepackage{amsmath}   

\newtheorem{theorem}{\bf Theorem}

\newtheorem{proposition}{\bf Proposition}
\newtheorem{lemma}{\bf Lemma}

\newtheorem{definition}{\bf Definition}

\ifCLASSINFOpdf
\else
\fi
\begin{document}
%
\title{The Structure Transfer Machine Theory and Applications}
%
%
%

\author{Wankou~Yang,~\IEEEmembership{Member,~IEEE,}
	    Baochang~Zhang,~\IEEEmembership{Member,~IEEE,}
        Ze~Wang, Lian Zhuo,
        Jungong~Han,~\IEEEmembership{Member,~IEEE,}
        Xiantong~Zhen,~\IEEEmembership{Member,~IEEE}

\thanks{ W. Yang is with the School of Automation, Southeast University, Key Laboratory of Measurement and Control of CSE, Ministry of Education, Nanjing 210096, China. E-mail: wkyang@seu.edu.cn}

\thanks{ B. Zhang Z. Wang, Z. Zhuo and X. Zhen are with Beihang University, Beijing, China. Baochang Zhang is also with Shenzhen
Academy of Aerospace Technology, Shenzhen 100083, China and  Key Laboratory of Measurement and Control of CSE, Ministry of Education, Nanjing 210096, China. Correspondence. E-mail: Bczhang@139.com}

\thanks{J. Han is with Department of Computer Science and Digital Technologies at Northumbria University, Newcastle, UK. E-mail: jungonghan77@gmail.com}

}

%
%

\markboth{IEEE Transactions on Image Processing,~Vol.~xx, No.~xx, xx~xxxx}%
{Shell \MakeLowercase{\textit{et al.}}: Bare Demo of IEEEtran.cls for IEEE Journals}
%



\maketitle

\begin{abstract}
Representation learning is a fundamental but challenging problem, especially when the distribution of data is unknown. In this paper, we propose a new representation learning method, named Structure Transfer Machine (STM), which enables feature learning process to converge at the representation expectation in a probabilistic way. We theoretically show that such an expected value of the representation (mean) is achievable if the manifold structure can be transferred from the data space to the feature space. The resulting structure regularization term, named manifold loss, is incorporated into the loss function of the typical deep learning pipeline. The STM architecture is constructed to enforce the learned deep representation to satisfy the intrinsic manifold structure from the data, which results in robust features that suit various application scenarios, such as digit recognition, image classification and object tracking. Compared with state-of-the-art CNN architectures, we achieve better results on several commonly used piblic benchmarks. 
\end{abstract}

\begin{IEEEkeywords}
Transfer learning, convolutional neural networks, manifold loss, learning theory.
\end{IEEEkeywords}

%
\IEEEpeerreviewmaketitle

\section{Introduction}


Human perception system abstracts the correct concept, when the relationship or the compactness of the intra-class data (small structure variations) is maintained after the perception; otherwise it will cause conceptual errors \cite{per}.  Analogously, data-driven learning approaches become a trend which by all means aim to maintain the class-specific feature compactness (perception) of the input data~\cite{J.Wright:PAMI2009, vgg, centerloss}. For a learning algorithm, such a compactness can be accomplished if the expected presentation is achieved for an unbiased estimator (classifier) \cite{bias1,bias2}. 
\begin{figure} [t!]
	\begin{center}
		\includegraphics[width=0.5\textwidth]{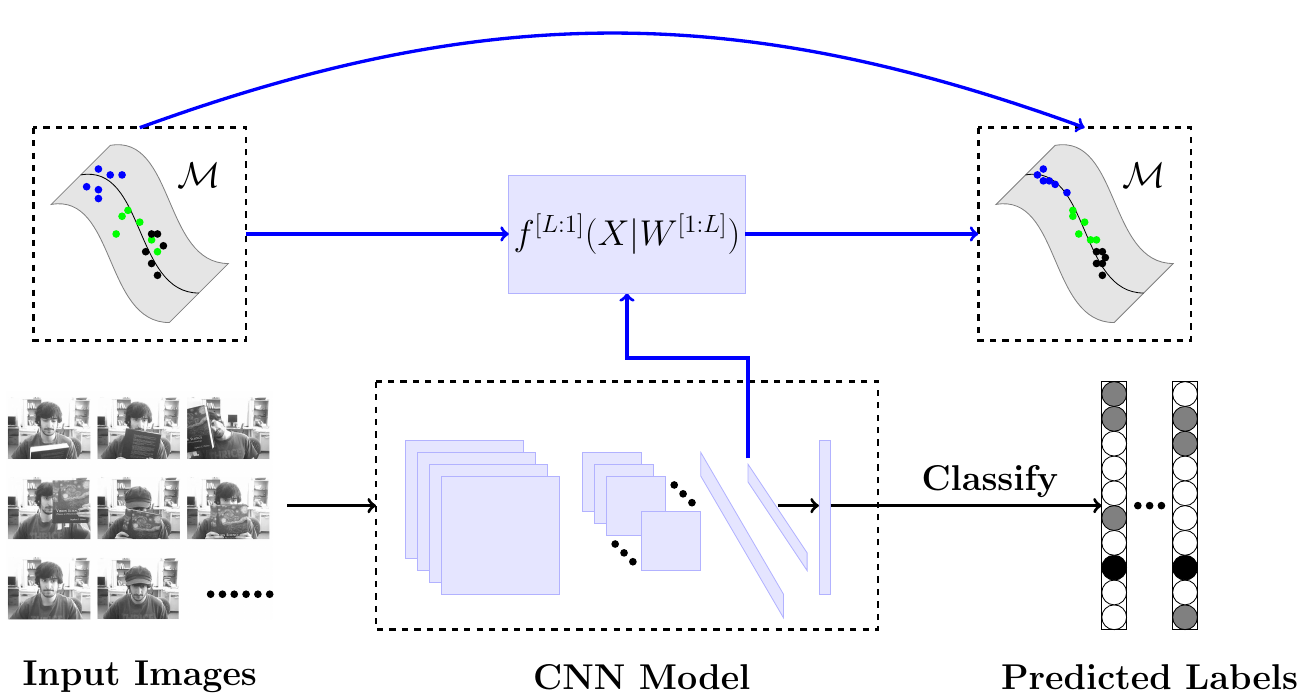}
	\end{center}
	\vspace{-0.1cm}
	\caption{Basic idea of the structure transferred machine (STM) method. By incorporating the manifold structure calculated in the input space into CNNs' loss function, termed as manifold loss, we can theoretically obtain the expected value of the representation (mean) in a probabilistic way, as a result the variations among local neighbors of the data are mitigated due to converging into the expected representation and thus gain the system robustness. Function $f$ is the feature mapping by CNNs (not necessarily the entire net) which will be introduced in section \ref{Sec.3.1}.}
	\vspace{-0.2cm}
	\label{fig.framework}
\end{figure}

Traditional hand-crafted features often require human expert knowledge, thereby making themselves domain specific. In contrast, deep learning based features can be learned automatically by composing multiple nonlinear transformations, yielding more abstract and useful representations. However, typically no distribution prior is embedded into the learning of deep features, making such schemes uncontrollable for certain circumstances. 
Recently, a center loss regularization term which is in essence a Gaussian prior is successfully exploited in deep learning to improve face recognition performance \cite{centerloss}. However, such a system does not work properly when the data is of complicated structure.  
Considering the fact that the conventional deep learning features are able to better distinguish the between-class variability~\cite{iclr,Lu2015CVPR}, we attempt to breakthrough the restriction of simple Gaussian prior \cite{centerloss} into a better prior so as to tolerate large intra-class variations. Thus, the inter-class samples can be still well separated even with the large intra-class variance due to super discriminant capability of deep learning. As a result, the generalization ability of the learned feature is expected to be significantly improved. 


In this paper, we discover that a desired representation in deep learning can actually be achieved and the local neighborhood with no constraint of the data structure required is able to converge at its expectation during the learning process. More importantly, it is noticed that the features describing the local structure enable representing the data better than the global ones, since the global features tend to be inaccurate when the data variation is usually large in real-world applications. The above observations inspire us to integrate a nonlinear manifold structure encoding more flexible structure of the data than the center loss \cite{centerloss} into the objective function of deep feature learning. Thus, we can accommodate variations among local neighbours such as rotations, rescalings, and translations so as to gain system robustness. However, directly embedding such data distribution into the deep learning framework is not an easy task at all, because formulating the underlying concept into appropriate training criteria is problematic. In this paper, we theoretically show that the expected representation can be achieved in a probabilistic way as long as the property of manifold structure is revealed in the objective function of deep learning. 

Upon such a proven, we present a novel structure transfer machine (STM) to learn structured deep features, the  framework of which is illustrated in Fig.~\ref{fig.framework}. Our STM starts with the incorporation of manifold structure into Convolutional Neural Network (CNNs) by calculating the manifold structure based on existing algorithms (i.e., local linear embedding (LLE) or Laplacian) in the input space. Such a manifold structure is in turn transferred to feature space and integrated into the loss function of the CNNs. Afterwards, the new CNN models are used to extract constitutional feature maps, where the intrinsic manifold structure is preserved even if it changes from the data space to the feature space. 
Experimental results demonstrate that these learned features can yield state-of-the-art performance in various computer vision tasks (e.g., digit recognition, natural object recognition, image classification (ImageNet) and object tracking) on commonly used benchmarks. Our main contributions include:

\begin{itemize}
	\item A theorem is developed to reveal that the expectation of representation can be obtained in a probabilistic way if a structure regularization is incorporated into the deep learning pipeline. With such structure regularization, we revise typical deep learning networks to a Structure Transfer Machine, which gains state-of-the-art performance on image classification and object tracking.
	
	\item With the aid of manifold, the structure of the input data is transferred into the feature space (output) with the intention to alleviate the unstructured problem in the higher-dimensional space, which eventually transfers the data structure into constraints in CNNs and leads to manifold loss. It is also demonstrated that in the deep feature space, the proposed manifold loss indeed improves the performance over the intra-class compactness methods such as the center loss.	
\end{itemize}

In the reminder of this paper, we analyze the related works in section \ref{Sec.2}. Subsequently, we propose the deep STM architecture and its corresponding theoretical analysis in section \ref{Sec.3}. In section~\ref{Sec.4}, we provide the experimental results as well as analysis. Finally, the last section draws conclusions.

\section{Related Work}\label{Sec.2}
Increasing the discriminating performance of the features learned by CNNs for images has received extensive attention in recent years. We roughly divide the related works into the following three parts.

\textcolor{black}{\textbf{Manifold learning.} Manifold learning methods \cite{tenenbaum2000a,Sam:Science,Laplacian,hettiarachchi2015multi,Saul:2003:TGF:945365.945372} assume that high dimensional data can be viewed as a set of geometrically related points lying on (or close to) the surface of a smooth low dimensional manifold. There are some manifold based learning methods published recently \cite{Le2011CVPR, iccvm2}, such as region manifold \cite{cvprm1}, graph manifold \cite{cvprm2}, product manifold \cite{cvprm3}, Grassmannia manifold \cite{cvprm5}, or its application to zero-shot \cite{cvprm4}. \textcolor{black}{We find that all of them are different from ours because the features in these methods are designed for specific tasks, such as region manifold exploring manifold for image retrieval, product manifold introducing the product manifold filter for the problem of bijective correspondence recovery, and Grassmannia manifold for spectral clustering. We focus on structure transferring in CNNs for general feature learning and actually provide a new theoretical investigation into CNNs.}}

\textcolor{black}{\textbf{Manifold regularization.} Highly relevant works contain the structure related regularization techniques~\cite{Lu2015CVPR} as well as the techniques that embed the prior knowledge, such as 2D topological structure of input data~\cite{Le2011CVPR}, both of which reveal that regularizing data structure is pretty useful when dealing with the image classification task. In \cite{iclr}, the adversarial examples suffer from performance degradation caused by small perturbations, manifold regularized networks (MRnet) that utilize a new training objective function aiming to minimize the difference between them. However, none of the existing works discuss the manifold constraint from theoretical perspective. Noted that in~\cite{Lu2015CVPR}, a manifold deep learning method is carried out for set classification. However, the difference between our work and their work is clear: we provide a theoretical investigation into the structure based deep learning, whereas the work in~\cite{Lu2015CVPR} is more empirical. From application perspective, we consider the intra-class information and focus on single image based classification, while ~\cite{Lu2015CVPR} is designed for set based classification by considering the inter-class information.}

\textcolor{black}{\textbf{Metric learning / Loss function.} Metric learning \cite{koestinger2012large,weinberger2009distance,ying2012distance} usually learns a matrix for a distance metric based on the given features. Recently, some state-of-the-art image classification (face recognition) models usually adopt ideas from metric learning. These methods \cite{Hu2014CVPR,Song2016CVPR,centerloss} use deep neural networks to automatically learn discriminative features followed by a simple distance metric such as Euclidean distance. For example, contrastive loss \cite{chopra2005learning,hadsell2006dimensionality} and triplet loss \cite{wang2014learning,hoffer2015deep,schroff2015facenet} are typical examples which borrow ideas from metric learning to increase the Euclidean margin for better feature embedding.  Center loss \cite{centerloss} forces CNNs to learn centers for the features of each label and uses the learned centers to reduce intra-class variance. Compared with the Euclidean margin (contrastive loss) or intra-class variance reduction (center loss), angular softmax (A-Softmax) loss (SphereFace) \cite{A-Softmax-loss}, ring loss \cite{ring-loss}, and cosine loss \cite{cosine-loss} implicitly involve the concept of angular margin. The angular margin is preferred because the cosine of the angle has intrinsic consistency with softmax. Our STM is different from these works: 1) the regularization item of deep features in these losses is predefined and independent of the structure of original data, while our manifold regularization item is dependent on the structure of original data; 2) the regularization item in these losses is defined on single data (image), while in our model it is defined on the distribution of all the original data, including both local information and global information.}


\section{Deep Structure Transfer Machine}\label{Sec.3}
It is reasonable to assume that the data lies on a manifold, whose intrinsic structure is expected to be embedded into the objective function of deep model. 
This is achieved by developing a generic representation learning method without any prior applied to the classification model. The proposed method is elaborated below, followed by the theoretical analysis.

\subsection{Problem formulation}\label{Sec.3.1}

\textcolor{black}{Let $D_{N}=\left\{(X_i, Y_i)|i=1,2,...,N\right\}$ be a training dataset, where $N$ is the total number of samples. 
\textcolor{black}{As shown in Fig.~(\ref{fig.framework}), we} define $\{F^{l}_{i}|l=1, 2, ..., L\}$ as a series of features for an image $X_i$ with the label $Y_i$ by a $L$-layers CNN, 
\begin{equation}
\begin{aligned}
X_i \longmapsto F^{1}_{i} \longmapsto F^{2}_{i} \longmapsto \cdots \longmapsto F^{L}_{i} \longmapsto \hat{Y}_i,
\label{eq.represents}
\end{aligned}
\end{equation}
where $\hat{Y}_i$ is the predicted label of $X_i$. The feature transfer functions of the CNN are defined as:
\begin{equation}
\begin{aligned}
F^{l}_{i} = f^{l}(F^{l-1}_{i}|W^{l}), \text{for}\ l=1, 2, ..., L+1,
\label{eq.feature.transfer}
\end{aligned}
\end{equation}
where $F^{0}_{i} = X_i$; $F^{L+1}_{i} = \hat{Y}_i$ and $W^{l}$ is the corresponding weight at $l$-th layer. Then, the relationship between feature $F^{l}_{i}$ and $X_{i}$ can be formulated as, 
\begin{equation}
\begin{aligned}
F^{l}_{i} = f^{l}(f^{l-1}(\cdots f^{1}(X_{i}|W^{1})\cdots|W^{l-1})|W^{l}).
\end{aligned}
\end{equation}
For the sake of simplicity, we have,
\begin{equation}
\begin{aligned}
F^{l}_{i} = f^{[l:1]}(X_{i}|W^{[1:l]}),
\end{aligned}
\end{equation}
where $l=1, 2, ..., L+1$. The transfer function of the output layer is usually a fully connected (FC) layer followed by a softmax function, and thus the loss function for the network is formulated as:
\begin{equation}
\begin{aligned}
{{\cal J}_{\lambda}}(W)
&= \frac{1}{N}\sum\limits_{i = 1}^N {{\cal L}\left(Y_i,\hat{Y}_i\right)}
+ \lambda\Omega(W)
\\
&= \frac{1}{N}\sum\limits_{i = 1}^N {-\log \frac{\exp(\theta_{{Y_i}}^{T}F^{L}_{i})}{\sum\nolimits_{j=1}^m {\exp(\theta_j^{T}F^{L}_{i})}}} + \lambda\Omega(W),
\label{eq.2}
\end{aligned}
\end{equation}
where $\theta=W^{L+1}$ denotes the weight matrix in the last FC layer; for simplicity we only use one FC layer as an example,
with $\theta_j$ as its $j$-th column; $m$ is the number of classes; the scalar $\lambda$ is a weight decay coefficient; and $\Omega(W)$ is a regularization function (e.g., $\Omega(W)=1/2\|W\|^2$) of all the the weights $W=\{W^{l}|l=1,2, ..., L+1\}$.} 

\begin{figure*}[h]
	\begin{center}
		\includegraphics[width=0.7\textwidth]{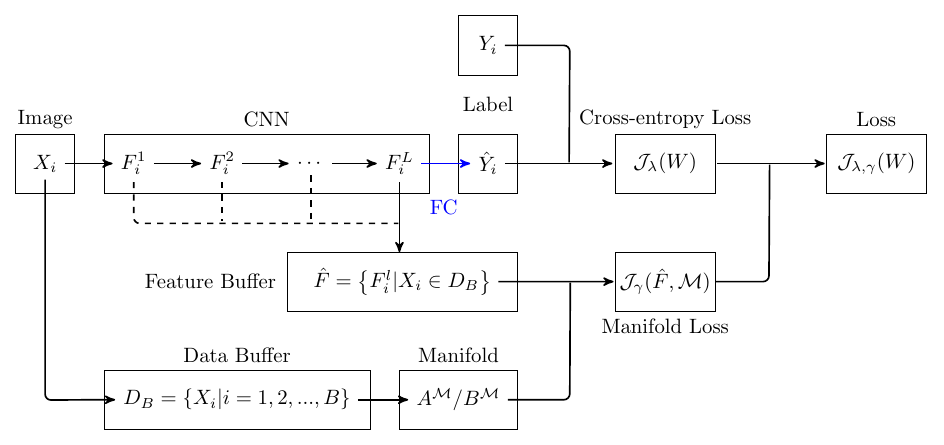}
	\end{center}
	\caption{\textcolor{black}{The architecture of STM.}}\label{architecture}
\end{figure*}

The conventional objective function for classification in Eq.~(\ref{eq.2}) does not consider the property that the data usually lies in a specific manifold ${\cal M}$~\cite{Zhang2015cvpr}, which reveals the nonlinear dependency of the data. Modeling this property can actually generate better solutions for lots of existing problems~\cite{Lu2015CVPR,iclr}. 
In the deep learning approach with error propagation from the top layer, it is more favorable to impose the manifold constraint on the top layer features. Our inspiration also comes from the idea of preserving manifold structure in different spaces, i.e., the high dimensional and the low dimensional spaces. Similarly, the manifold structure of ${X}$ is assumed to be preserved in the resulting deep features \textcolor{black}{$F^l$} of our model in order to reduce variation in the higher-dimensional feature space (\textcolor{black}{Fig.~\ref{fig.framework}}). 
We resort to a new manifold constraint in deep learning, and achieve a \textcolor{black}{new problem~(\ref{P2}),}
\textcolor{black}{
\begin{equation*}
\begin{aligned}
{{\cal J}_{\lambda}}(W) &= \frac{1}{N}\sum\limits_{i = 1}^N {{\cal L}\left(Y_i, \hat{Y}_i\right)} + \lambda\Omega(W) \\
\text{s.t.} & \quad {\hat{F}_i \in {\cal M}}, \quad \text{for\ } i=1, 2, ..., N,
\end{aligned}
\tag{$P_1$}
\label{P2}
\end{equation*}
where $\hat{F}_i$ can be the deep feature of any layer, i.e., $\hat{F}_i = F_i^{l} $, $l \in \{1, 2, ..., L\}$. 
It is noticed that the objective shown in problem~(\ref{P2}) is learnable if ${\cal M}$ is given, because $ \hat{F}_{i}$ is directly related to the learned filters (Eq.~(\ref{eq.represents}) and Eq.~(\ref{eq.feature.transfer})). How to solve the constraint $ \hat{F}_i \in {\cal M}$  is elaborated in the next section.}

\subsection{Manifold loss}\label{Sec.3.3}

Solving the above problem needs to know manifold $\cal{M}$. Here we hypothesize it to be any manifold, e.g., LLE and Laplacian. 

\textbf{LLE.} According to LLE, it is assumed that each data point and its neighbors lie on a locally linear patch of the manifold. Hence we compute the linear coefficients \textcolor{black}{$A^{{\cal M}}$} to reconstruct each data from its neighbors by minimizing the reconstruction error:
\begin{equation}
\label{reconstruction}\textcolor{black}{
\begin{aligned}
\varepsilon(A^{\mathcal{M}}) &= \sum_{i=1}^{N}{\|X_i-\sum_{X_j\in{k\text{-NN}(X_i)}}{\alpha_{ij} X_j}\|^2}\\
&= \sum_{i=1}^{N}{\|X_i - X  A^{{\cal M}}_{i}\|^2},
\end{aligned}}
\end{equation}
which is a manifold loss. Here we define \textcolor{black}{$A^{\cal M}_{i} = {[{\alpha}^{}_{i1}, {\alpha}^{}_{i2}, ... ,{\alpha}^{}_{iN}]}^T$} with \textcolor{black}{${\alpha}_{ij}$} being the corresponding weights of neighborhood data \textcolor{black}{$X$,} which is actually the feature buffer set, \textcolor{black}{as shown in Fig.~\ref{architecture}}, for the \textcolor{black}{$i$-th} data $X_i$ in the original data space. We enforce \textcolor{black}{${\alpha}^{}_{ij}=0$} if $ X_j$ does not belong to the neighborhood of $X_i$ \textcolor{black}{(i.e., $X_j\notin{k\text{-NN}(X_i)}$)}, such that each point is only reconstructed by its neighbors. The optimal weights \textcolor{black}{$A^{{\cal M}}$} can be found by solving a least square problem with constraint \textcolor{black}{$\sum_j{\alpha_{ij}}=1$}. As assumed, a linear embedding process for neighborhood preserving in the feature space is given by:
\begin{equation}\label{bczhangm}
\textcolor{black}{\hat{F}_{i} = \hat{F}A^{{\cal M}}_{i},}
\end{equation}
where \textcolor{black}{$\hat{F}_{i}^{}$} is the deep feature from the current layer for the $i$-th input sample, and the feature of its neighbors or feature buffer \textcolor{black}{(as shown in Fig.~\ref{architecture})} are denoted by  \textcolor{black}{$\hat{F}$}. In this process, the feature of each sample is linearly reconstructed from \textcolor{black}{$\hat{F}$} by linear coefficients. The reconstruction weight \textcolor{black}{$A^{\cal M}$} is obtained by minimizing \textcolor{black}{Eq.~(\ref{reconstruction})}, which characterizes intrinsic geometric properties of the data that are invariant to rotations,
rescalings, and translations of that data point and its neighbors \cite{Sam:Science}, is related to the manifold $\cal M$. This is the key part of the proposed algorithm where the constraint manifold ${\cal M}$ arises. As assumed, replacing ${\cal M}$ equals incorporating \textcolor{black}{Eq.~(\ref{bczhangm})} into our objective. This is the modularity alluded previously. Based on the Lagrangian multiplier method, \textcolor{black}{Eq.~(\ref{bczhangm})} is introduced to solve problem~(\textcolor{black}{\ref{P2}}) by a new objective as:
\begin{equation}\label{lle}
\textcolor{black}{
\begin{aligned}
{{\cal J}_{\lambda, \gamma}}(W) &= {{\cal J}_\lambda }(W) + \mathcal{J}_{\gamma}(\hat{F}, \mathcal{M})\\
&= {{\cal J}_\lambda }(W) + \frac{\gamma}{2N} \sum_{i = 1}^N \| {\hat{F}_i - \hat{F}  A^{\cal M}_{i}}\|^2,
\end{aligned}}
\end{equation}
where the scalar $\gamma$ is adopted to balance these two terms of the objective.

\textbf{Laplacian.} Similar to LLE, we can also exploit the Laplacian manifold~\cite{Laplacian} to our problem. As shown in~\cite{Laplacian}, we can deduce such a manifold loss:
\begin{equation}\textcolor{black}{
\mathcal{J}_{\gamma}(\hat{F}, \mathcal{M}) = \sum_{i=1}^{N} \sum_{j=1}^{N} \|\hat{F}_{i} - \hat{F}_{j}\|^2 {B^{\cal M}_{ij}},}
\end{equation}
where $B^{\cal M}_{ij}$ is defined to be exponential distance between the $i$-th and $j$-th sample in the input space~\cite{Laplacian}, \textcolor{black}{which is actually used to normalize the feature buffer set}\textcolor{black}{, as shown in Fig. \ref{architecture},} to improve the efficiency. Similarly, we obtain the following objective as:
\begin{equation}\label{laplacian}\textcolor{black}{
\begin{aligned}
{{\cal J}_{\lambda, \gamma}}(W) & = {{\cal J}_\lambda }(W) + \mathcal{J}_{\gamma}(\hat{F}, \mathcal{M})\\
& = {{\cal J}_\lambda }(W) + \frac{\gamma}{2N^2} \sum_{i=1}^N \sum_{j=1}^N \|\hat{F}_i^{} - \hat{F}_j^{}\|^2 B^{\cal M}_{ij}.
\end{aligned}}
\end{equation}

\subsection{Training algorithm}\label{sec-training}
\textcolor{black}{Now, we have the training algorithm of STM to solve problem~(\ref{P2}), which is summarized in \textbf{Algorithm~\ref{stm}}.  Regarding the convergence of the proposed algorithm, our learning procedures never hurt the convergence of the back propagation, because newly added variables related to the manifold loss (convex) are solved following the similar pipeline. As shown in \textbf{Algorithm~\ref{stm}}, the procedures for training STM are as follows.}

\textcolor{black}{The input of the training algorithm includes a set of image-label pairs $D_N = \{(X_i, Y_i)|i = 1, 2, ..., N\}$, the number of mini-batch size $N_b$, the learning rate $\mu$, the size of the feature buffer size $k_0$ as shown in Fig.~\ref{architecture}, the number of neighbors in LLE $k_l$, the weight decay parameter $\lambda$, the manifold loss balance parameter $\gamma$, the indicator $m\in\{{lle}, {lap}\}$ for denoting which manifold method is used in training, the maximum number of iterations $\tau_{\max}$, and the minimum value of {the objection function} $\mathcal{J}_{\min}$. The output of this algorithm is the trainable parameter $W$.}

\textcolor{black}{\textbf{Step 1. Initialization, line 1.} Initialize $t$ with $0$, $D_{B}$ with $\phi$, $\hat{F}$ with $\phi$ and ${W}$ with random values \cite{pmlr-v9-glorot10a,7410480,pmlr-v37-ioffe15}, where $\phi$ means empty set.} 

\textcolor{black}{\textbf{Step 2. Picking up mini-batch samples, line 3.} Randomly pick $N_b$ input-output pairs $(X_i, Y_i)$ as a mini-batch sample set $D_{N_b}$. The time complexity and space complexity of this step are both equal to the sample complexity, i.e.,  $O(N_b\times |D|)$, where $|D| = |X|+|Y|$ denotes the capacity of one input-output pair. }

\begin{algorithm}[h]
	\SetKwInput{Input}{Input}
	\SetKwInput{Output}{Output}
		
	\Input{\textcolor{black}{$D_{N}$, $N_b$, $u$, $k_0$, $k_l$, $\lambda$, $\gamma$, ${m}$, $\tau_{\max}$, $\mathcal{J}_{\min}$;}}
	\Output{\textcolor{black}{$W$;}}
	
	\textcolor{black}{Initialize $t$, $D_{B}$, $\hat{F}$ and ${W}$ with default or random values\;}
	
	\Repeat{\textcolor{black}{$t>\tau_{\max}$ $\lor$ $\mathcal{J}_{\lambda, \gamma}(W) < \mathcal{J}_{\min}$}}
	{
		\textcolor{black}{$D_{N_b}\sim \mathcal{G}(D_N)$\;}
		\textcolor{black}{
		\For{$X_i \in D_{N_b}$}{
			\For{$l\in\{1, 2, ..., L+1\}$}{
				$F^{l}_{i} \gets f^{[l:1]}(X_{i}|W^{[1:l]})$\;
			}
		}}
		
		\textcolor{black}{$\{D_{B}, \hat{F}\} \gets \text{\textit{refresh}}(D_{B}, \hat{F}, D_{N_b}, \{F^{L}_{i}|X_i\in D_{N_b}\}, \gamma, k_l, m)$\;}
			
		\textcolor{black}{${{\cal J}_{\lambda}}(W) \gets \frac{1}{N_{b}}\sum\limits_{i = 1}^{N_b} {{\cal L}\left(Y_i,\hat{Y}_i\right)} + \lambda\Omega(W)$\;}

		\textcolor{black}{$\mathcal{J}_{\gamma}(\hat{F}, \mathcal{M}) \gets m(D_B, \hat{F}, \gamma, k_l)$\;}
	
		\textcolor{black}{${{\cal J}_{\lambda, \gamma}}(W) \gets {{\cal J}_\lambda }(W) + \mathcal{J}_{\gamma}(\hat{F}, \mathcal{M})$\;}
		
		\textcolor{black}{$\nabla_{W} \gets \frac{\partial {{\cal J}_{\lambda}}(W)}{ \partial W} + \frac{\partial \hat{F}}{\partial W}\frac{\partial \mathcal{J}_{\gamma}(\hat{F}, \mathcal{M})}{\partial \hat{F}}$\;}
		
%
		\textcolor{black}{$W \gets W - \mu \nabla_{W}$\;}
		
		\textcolor{black}{$t \gets t + 1 $\;}	
	}

	\textcolor{black}{\Return  $W$\;}
	
	\caption{\textcolor{black}{Training algorithm of STM.}}\label{stm}
\end{algorithm}

\textcolor{black}{\textbf{Step 3. Forward-propagation (i.e., calculating features and labels for the mimi-batch samples), line 4-8.} The space complexity of this step is $O(1)$, and the time complexity of this step is $O(N_b\times|W|)$.}

\textcolor{black}{\textbf{Step 4. Refreshing the data buffer and feature buffer, line 9.} The details of function $refresh(\cdot)$ and its corresponding complexities are shown in \textbf{Algorithm~\ref{algo-refresh}} in Appendix~\ref{app-algorithm-refresh}. We denote the time complexity and space complexity of $refresh(\cdot)$ by $\mathcal{T}_{r}$ and $\mathcal{S}_{r}$.}

\textcolor{black}{\textbf{Step 5. Calculation of the objective function, line 10-12.} The total value of the objective function (line 12) consists of two items: ordinary loss (line 10) and manifold loss (line 11). The time complexity of this step is $O(|W|)+\mathcal{T}_{lle}$ if we using LLE manifold, or $O(|W|)+\mathcal{T}_{lap}$ if Laplacian manifold is used; and the space complexity is $\mathcal{S}_{lle}$ or $\mathcal{S}_{lap}$. The $\mathcal{T}_{lle}$, $\mathcal{S}_{lle}$, $\mathcal{T}_{lap}$ and $\mathcal{S}_{lap}$ are used to denote the time complexity, space complexity of function $lle(\cdot)$ and function $lap(\cdot)$. The details for calculating these functions are shown in \textbf{Algorithm~\ref{algo-lle}} in Appendix~\ref{app-algorithm-lle}, and in \textbf{Algorithm~\ref{algo-lap}} in Appendix~\ref{app-algorithm-lap}.}

\textcolor{black}{\textbf{Step 6. Back-propagation (i.e., calculating the gradient), line 13.} Calculate the gradient $\nabla_{W}$ with back-propagation. The space complexity and time complexity of this step are $O(1)$ and $O(|W|)$, respectively.}

\textcolor{black}{\textbf{Step 7. Updating the trainable parameters, line 14.} The time complexity of this step is $O(|W|)$; and the space complexity of this step is $O(1)$.}

\textcolor{black}{\textbf{Step 8. Not reaching the end condition, line 15-16.} Increase the number of iterations $t\leftarrow t+1$. If $t \le \tau_{\max}$ and $\mathcal{J}_{\lambda, \gamma}(W) \ge \mathcal{J}_{\min}$, go to \textbf{Step 2}.} 

\textcolor{black}{\textbf{Step 9. Return the output, line 17.} Return the trained parameters $W$.}

\textcolor{black}{In summary, the time complexity of the training algorithm of STM in each mini-batch training is}
\textcolor{black}{\begin{align}\label{eq-time-comlexity}
\mathcal{T} = O(N_b\times(|W|+ |D|)+3|W|)+\mathcal{T}_{r}+\mathcal{T}_{m},
\end{align}}
\textcolor{black}{and the space complexity in each mini-batch training is}
\textcolor{black}{\begin{align}\label{eq-space-comlexity}
\mathcal{S} = O(N_b\times(|W|+ |D|))+\mathcal{S}_{r}+\mathcal{S}_{m},
\end{align}}\textcolor{black}{where $m\in\{{lle}, {lap}\}$ is used for denoting which manifold loss function is used. From Eq.~(\ref{eq-time-comlexity}) and Eq.~(\ref{eq-space-comlexity}), we know that the last two items are the additional complexities which come from two parts: the function of refreshing the data buffer and features buffer, and the function of calculating  manifold loss.}

\subsection{Theoretical analysis} 
In this section, we theoretically show that the manifold loss can lead a convergence process to the expectation of data representation, based on assumption that data lies on a manifold. More specifically, \textcolor{black}{\textbf{Theorem~\ref{theorem-2}}} provides a foundation of feature learning that the expected value can be achieved, if manifold structure is transferred from the input space to the feature space. Such a proof is very useful to guide the feature design in various  practical applications. Notably, many machine learning tasks often require that features are compact and stable during the model learning process. In other words, the variances among the data are mitigated in the learning process, as they are converging into a single expected value. In the following, we will address how our theorem can be involved in the learning stage. 




\begin{definition}\label{definition-1}
	For  $x_1,x_2,...,x_n$, define: 
	\begin{equation}
	\|x_{i-1}-x_{i}\| \le c_{i-1} ,
	\end{equation}
	where $x_i$ is a random variable and $c = (c_0, c_1,..., c_{n-1})$. \textcolor{black}{Further we have $ x_{i} = x_{i-1} + c_{i-1}$, if define $$c_{i-1} \sim \mathcal{N}(0,\sigma^2),$$ where $\mathcal{N}$ is a Gaussian distribution with $0$ mean and $\sigma$ standard variation.} 
\end{definition}

\textcolor{black}{From the above definition, we know that $x_{i}$ and $x_{i-1}$ are bounded by $c_{i-1}$. The expectation of $x_{i}$ given $x_{i-1}$ is chosen to be $x_{i-1}$, which means that $x_{i}$  is sampling from a distribution that is quite related to  $x_{i-1}$. That is to say, $x_{i}$ is chosen around $x_{i-1}$ (mean).}


\begin{lemma}\label{theorem-1}
	If $x_{1},x_{2},...,x_{n}$ satisfies \textcolor{black}{\textbf{Definition~\ref{definition-1}}}, then: 
	\begin{equation}\label{eq-theo-1}\textcolor{black}{
	P(|\bar{x}-E(\bar{x})| \ge \frac{\lambda}{n}) \le 2\exp(\frac{-\lambda^2}{2\sum_{i=1}^{n}{c_i^2}}),}
	\end{equation}
	where $\bar{x}$ is the average of $x_{1},x_{2},...,x_{n}$.
\end{lemma}
\begin{IEEEproof}
	\textcolor{black}{The details are shown in Appendix \ref{app-proof-lemma-1}.}
\end{IEEEproof}

Before proving the theorem, we first introduce the following two propositions.

\begin{proposition}\label{proposition-1}
	The most popular approaches, such as LLE \cite{Sam:Science} and ISOMAP \cite{tenenbaum2000a}, are with the underlying idea that a high dimensional vector representing the data that can be mapped into a lower dimension space preserving, as much as possible, the metric of the original space. The distances of all the pairs of data points in the embedding space is bounded \cite{lips1,lips3}. Thus, it is claimed that:
	\begin{equation}\label{claim}\textcolor{black}{
	\|{F}_{i}(j)- {F}_{i-1}(j)\| \le \|{F}_{i}- {F}_{i-1}\| \le L \|X_{i-1}^{}-X_{i}^{}\|,}
	\end{equation} \textcolor{black}{where ${F}_{i} = f(X_{i})$, and $f(\cdot)$ denotes the projection from the original sample $X_{i}$ to ${F}_{i}$. And ${F}_{i}(j)$ denotes the $j$-th dimension of $F_i$ in the manifold feature space, i.e., the deep feature space obtained based on manifold loss in this work; $L$ is a constant. Due to  $\|X_{i}-X_{i-1}\|$ is controlled by the input sample, so that it is reasonable to claim that ${F}_{i}$:}
	\begin{equation} \label{claim1}\textcolor{black}{
	\|{F}_{i}(j) - {F}_{i-1}(j)\| \le \bm{c}_{i-1}(j),}
	\end{equation}
	where, $\bm{c}_{i}(j)$ is the $j$-th dimension of vector $\bm{c}_i$.
\end{proposition}
\begin{proposition}\label{proposition-2}\textcolor{black}{
	For any vector $v=[v_1, v_2, ..., v_n]$, we have:}
	\begin{equation}\label{k12}\textcolor{blue}{
	\begin{aligned}
	P(\sum_{i=1}^{n}{|v_{i}|} \ge \sum_{i=1}^{}{\lambda_{i}}) &\le P(\bigcup_{i=1}^{n}{|v_{i}| \ge \lambda_{i}})\\
	& \le \sum_{i}^{n}{P(|v_{i}| \ge \lambda_{i})}.
	\end{aligned}}
	\end{equation}
\end{proposition}

Now, we have the following theorem. 
\textcolor{black}{
	\begin{theorem}\label{theorem-2}
		If vectors \textcolor{black}{${F}_1,{F}_2,...,{F}_n \in {\cal M}$}, then:
		\begin{equation}\textcolor{black}{
				P(|\bar{F}-E(\bar{F})| \ge \sum_i \lambda_i) \le C,}
		\end{equation} where  $\bar{F}$ is the average vector; $ \mathbb{\lambda} = [\lambda_1,\lambda_2,...,\lambda_n]$ with $\lambda_i \le 1$ and $C$ is a constant.
\end{theorem}}

\begin{IEEEproof}
\textcolor{black}{\textbf{Theorem~\ref{theorem-2}}} means that the expectation of \textcolor{black}{$\bar{F}$} is achieved in a probabilistic way. According to \textcolor{black}{Eq.~(\ref{eq-theo-1}) in \textbf{Lemma~\ref{theorem-1}} and Eq.~(\ref{claim1}) in \textbf{Proposition~\ref{proposition-1}}}, we have:
\begin{equation}\textcolor{black}{
P(|Z(j)| \ge \lambda_{j}) \le 2\exp(\frac{- \lambda^{2}_{j}}{2 \sum_{i=1}^{n}{\bm{c}^{2}_{i-1}(j)}}),}
\end{equation}
where \textcolor{black}{$Z$ is defined as:}
\begin{equation}\textcolor{black}{
	Z=\bar{F}-E(\bar{F}).} 
\end{equation} 
We set \textcolor{black}{$a_j = 2\exp(- \lambda^{2}_{j} / (2 \sum_{i=1}^{n}{\bm{c}^{2}_{i-1}(j)})$}, based on \textcolor{black}{Eq.~(\ref{k12}) in \textbf{Proposition~\ref{proposition-2}}}, we have:
\begin{equation}\textcolor{black}{
P({|\bar{F}-E(\bar{F})|} \ge \sum_{j=1}^{n}{\lambda_{j}}) \le \sum_{j=1}^{n}{a_j}=C.}
\end{equation}

Thus, \textcolor{black}{\textbf{Theorem~\ref{theorem-2}}} is proved.
\end{IEEEproof}

\section{Experiment}\label{Sec.4}
In this section, we first present the details about how to implement our method with a deep learning pipeline. \textcolor{black}{We then use the digit recognition (MNIST) and the natural object recognition (CIFAR) experiments to show the superiority of our method.} We finally validate the effectiveness of our method with large-scale visual tasks including image classification  and object tracking.

\subsection{Implementation details}
\textbf{Comparison.} We validate our method on various CNN base models, including ResNet \cite{rsn}, WideResNet \cite{cifar_rsn}, and then compare the performance with state-of-the-art networks. \textcolor{black}{Center loss \cite{centerloss}, A-Softmax loss (SphereFace) \cite{A-Softmax-loss}, ring loss \cite{ring-loss}, and cosine loss \cite{cosine-loss} are also evaluated equally as comparison.} For the unavailability of the training face database used in \cite{centerloss}, we choose other testbeds, such as MNIST, CIFAR, ImageNet and the large scale OTB-50 tracking database for a fair comparison. 

\textbf{Manifold.} We introduce manifold loss or the constraint term for structure preserving based on LLE or Laplacian. 
\textcolor{black}{We build a feature buffer (\textcolor{black}{Fig. \ref{architecture}}) consisting of its $k_0$ (e.g., 30) previous samples from the same class, which denotes the maximum number of nearest neighbors used to calculate the reconstruction weights ($A^{{\cal M}}$, $B^{{\cal M}}$) exactly as that in LLE or Laplacian manifold.} 
Notice that we then obtain the feature from this sample mapped by the network in each iteration and its corresponding subset of features from its neighbor samples, and learn the model by the proposed loss function as the reconstruction weights (\textcolor{black}{$A^{{\cal M}}$, $B^{{\cal M}}$}) are introduced. The above process is used for the MNIST, CIFAR and ImageNet datasets, but in the tracking task we divide each sequence into batch sets, which are then used to calculate the manifold for further learning process.

\textbf{Settings in CNN.} The proposed models are implemented on common libraries (i.e., \textcolor{black}{Caffe, TensorFlow and PyTorch}) with our modifications, and can still be trained end-to-end by SGD without introducing many parameters compared to their base model. We train our STMs via the algorithm shown in \textcolor{black}{section~\ref{sec-training}}. The manifold loss is added before the FC layer as shown in \textcolor{black}{Fig.~\ref{architecture}}. To further understand how the elements of the framework affect the performance, we test our models when the manifold data structure is extracted by different techniques, such as \textcolor{black}{LLE or Laplacian}. 
For fair comparison, most of settings in our networks follow the ones in their base model, except for the learning rate and policy, because the modified object function is determined on validation set. More detailed settings in each experiment are described in corresponding subsection.

\subsection{Digit recognition}
\textbf{MNIST} dataset of handwritten digits\footnote{http://yann.lecun.com/exdb/mnist/} contains a training set of 60,000 examples, and a test set of 10,000 examples. It is a subset of a larger set available from NIST. The digits have been size-normalized and centered in a fixed-size image.

We use a weight decay \textcolor{black}{($\lambda$)} of 0.0001 and momentum of 0.9 in our model with a mini-batch size \textcolor{black}{($N_b$)} of 500. The learning rate \textcolor{black}{($\mu$)} is started from 0.1, and divided by 10 at 32k and 48k iterations, and the training procedure is terminated at 64k iterations. We do not conduct any data augmentation for training. The \textcolor{black}{LeNet++ \cite{centerloss} architecture is used for base-CNN, our STM, center loss \cite{centerloss}, A-Softmax loss \cite{A-Softmax-loss}, ring loss \cite{ring-loss}, and cosine loss \cite{cosine-loss}} for a fair comparison.

\begin{table}[h]
	\caption{\textcolor{black}{Comparisons with different hyper-parameters of STM on the MNIST dataset.}}
	\label{parameters}
	\centering
	\begin{tabular}{|L{2.35cm}|C{0.7cm}|C{0.7cm}|C{0.7cm}|C{1.35cm}|}
		\hline
		\multirow{2}{*}{\textbf{Models}} & \multicolumn{3}{c|}{\textbf{Hyper-parameter setting}} & \multirow{2}{*}{\textbf{Error (\%)}}\\
		\cline{2-4} & \textbf{$k_l$} & \textbf{$k_0$} & \textbf{$\gamma$} & \\
		\hline
		\hline
		\textcolor{black}{\multirow{17}{*}{STM with LLE}} & 14 & 30 & 0.000 & 0.73\\
		& 14 & 30 & 0.001 & 0.65\\
		& 14 & 30 & 0.010 & 0.45\\
		& 14 & 30 & 0.100 & 0.40\\
		& 14 & 30 & 0.200 & 0.37\\
		& 14 & 30 & 0.300 & \textbf{0.36}\\
		& 14 & 30 & 0.400 & \textbf{0.36}\\
		& 14 & 30 & 0.500 & 0.37\\
		& 14 & 30 & 0.800 & 0.37\\
		& 14 & 30 & 0.900 & 0.38\\
		& 14 & 30 & 1.000 & 0.38\\
		& 22 & 30 & 0.000 & 0.73\\
		& 22 & 30 & 0.100 & 0.45\\
		& 22 & 30 & 0.200 & 0.42\\
		& 22 & 30 & 0.300 & 0.41\\
		& 22 & 30 & 0.800 & 0.42\\
		& 22 & 30 & 1.000 & 0.42\\
		\hline		
		\textcolor{black}{\multirow{12}{*}{STM with Laplacian}} 
		& 14 & 30 & 0.000 & 0.73\\
		& 14 & 30 & 0.001 & 0.60\\
		& 14 & 30 & 0.010 & 0.41\\
		& 14 & 30 & 0.100 & 0.38\\
		& 14 & 30 & 0.200 & \textbf{0.36}\\
		& 14 & 30 & 0.300 & \textbf{0.36}\\
		& 14 & 30 & 0.400 & \textbf{0.36}\\
		& 14 & 30 & 0.800 & 0.38\\
		& 14 & 30 & 1.000 & 0.38\\
		& 22 & 20 & 0.200 & 0.38\\
		& 22 & 20 & 0.300 & 0.37\\
		& 22 & 20 & 0.400 & 0.38\\
		\hline
	\end{tabular}
\end{table}

\textbf{(1) Parameter evaluation and performance comparison.} There are several parameters affecting the performance of the proposed method, i.e., \textcolor{black}{$k_0$} denoting the size of the feature buffer set for each class, \textcolor{black}{$k_{l}$} denoting number of neighbors in LLE or Laplacian. The results in \textcolor{black}{TABLE~\ref{parameters}} show that STM with LLE achieves the best performance when \textcolor{black}{$k_{l} = 14$}. \textcolor{black}{The performances of LLE and Laplacian based STMs are very similar, but STM with LLE needs more computation as shown in Algorithm~\ref{algo-lle}, in comparison with STM with Laplacian (Algorithm~\ref{algo-lap}). For all the following experiments the buffer size and neighbor size are set as, \textcolor{black}{$k_0=30$, $k_l=14$}. The parameter \textcolor{black}{$\gamma$} is also evaluated \textcolor{black}{ in TABLE~\ref{parameters}}, which show that in the certain scope the parameter affect little on the final performance.}

\begin{figure*}[h]
	\centering
	\begin{minipage}{0.13\textwidth}
		\centering
		\subfigure[]{
			\centering
			\includegraphics[height=2.35cm]{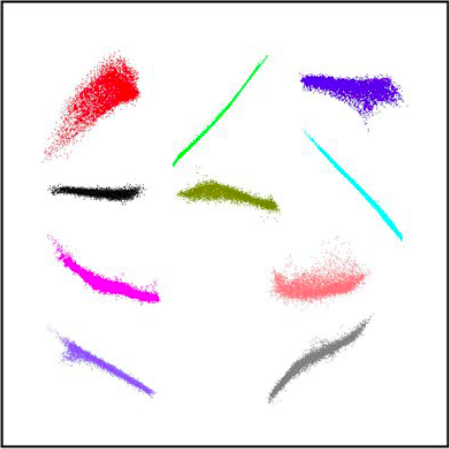}
		}
	\end{minipage}	
	\begin{minipage}{0.13\textwidth}
		\centering
		\subfigure[]{
			\centering
			\includegraphics[height=2.35cm]{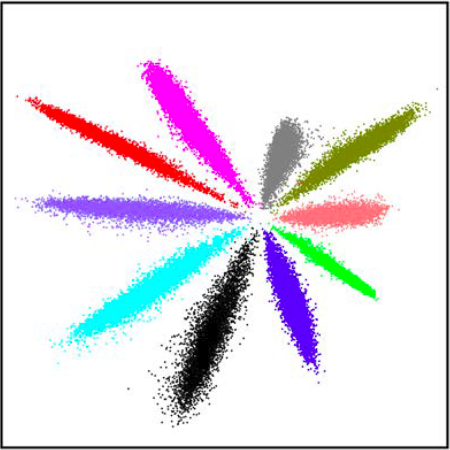}
		}
	\end{minipage}	
	\begin{minipage}{0.13\textwidth}
		\centering
		\subfigure[]{
			\centering
			\includegraphics[height=2.35cm]{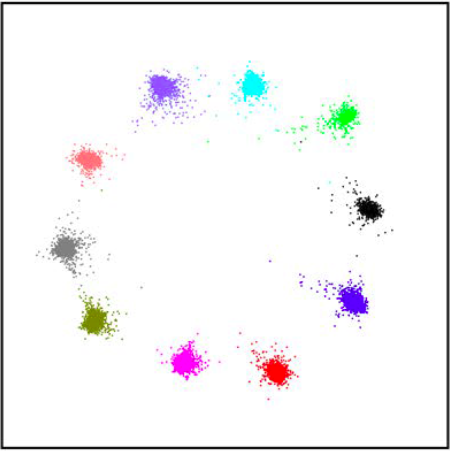}
		}
	\end{minipage}	
	\begin{minipage}{0.13\textwidth}
		\centering
		\subfigure[]{
			\centering
			\includegraphics[height=2.35cm]{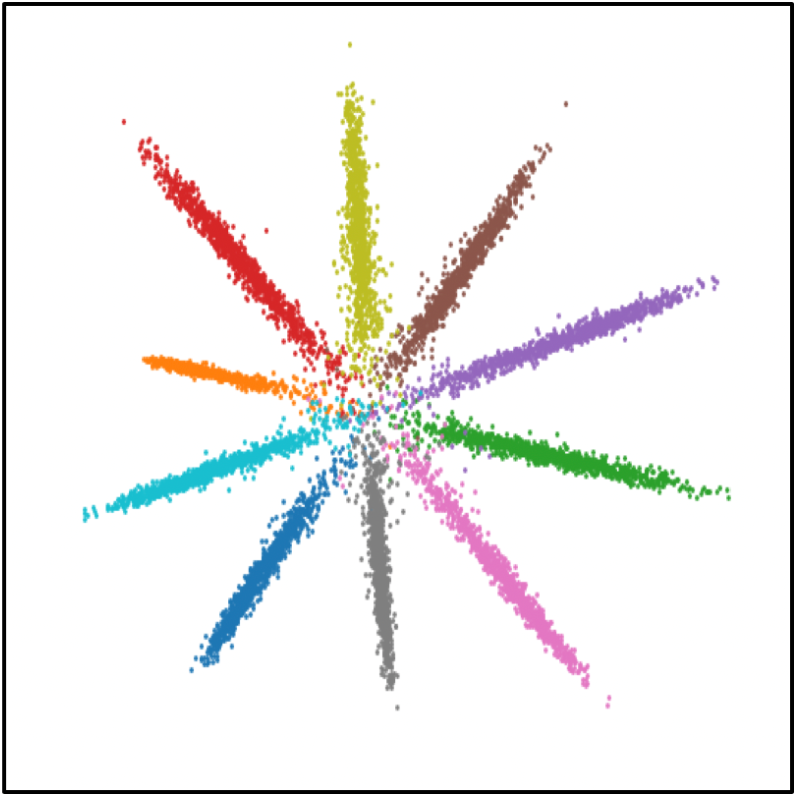}
		}
	\end{minipage} 
	\begin{minipage}{0.13\textwidth}
		\centering
		\subfigure[]{
			\centering
			\includegraphics[height=2.35cm]{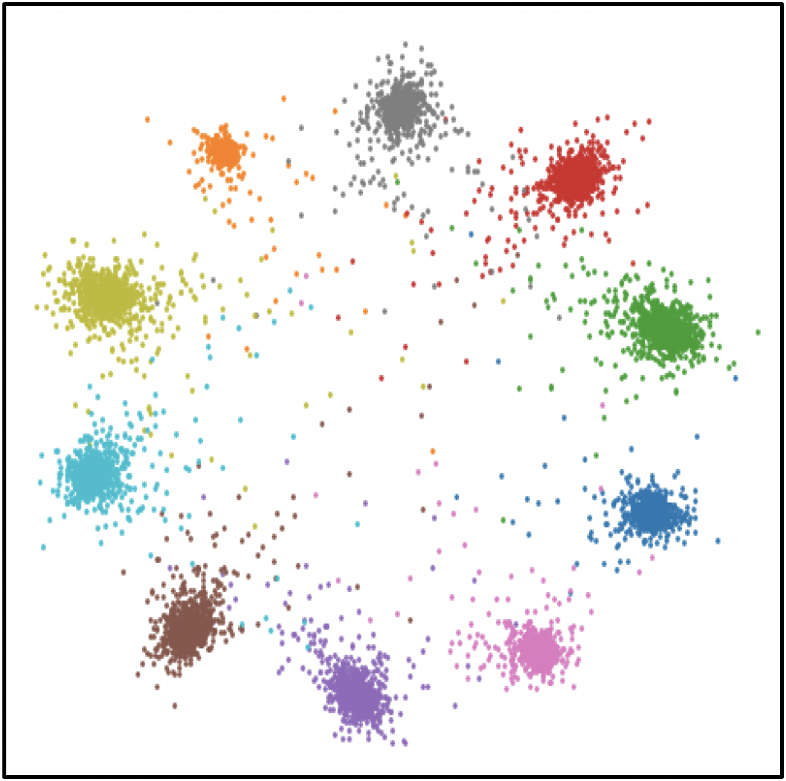}
		}
	\end{minipage}	
	\begin{minipage}{0.13\textwidth}
		\centering
		\subfigure[]{
			\centering
			\includegraphics[height=2.35cm]{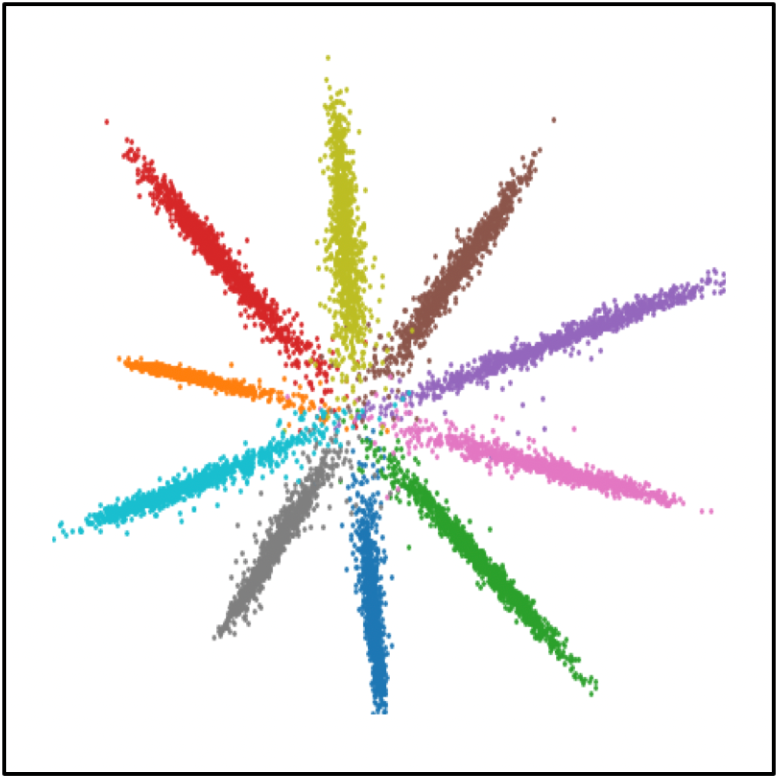}
		}
	\end{minipage}
	\begin{minipage}{0.13\textwidth}
		\centering
		\subfigure[]{
			\centering
			\includegraphics[height=2.35cm]{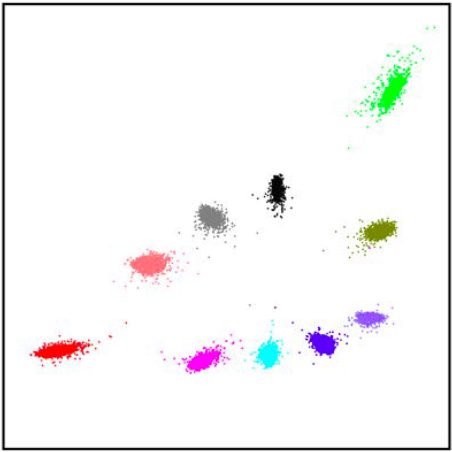}
		}
	\end{minipage}
	\caption{Distribution illustration of learned features, where (a) is the manifold, and we create each structure separately but show in one figure;  (b) is baseline CNN feature; \textcolor{black}{(c) is center loss \cite{centerloss} feature; (d) is A-Softmax loss (SphereFace) \cite{A-Softmax-loss} feature; (e) is ring loss \cite{ring-loss} feature; (f) is cosine loss \cite{cosine-loss} feature;} and (g) is the STM feature.}
	\label{illustration}
\end{figure*}

\textbf{(2) Illustration.} We first conduct experiments to illustrate how the STM method influences the distribution in \textcolor{black}{Fig.~\ref{illustration}}. Without the special note, STM means using Laplacian to calculate \textcolor{black}{$B^{\cal M}$} from the original data space for learning. \textcolor{black}{\textcolor{black}{Fig.~\ref{illustration}} shows that the distributions of STM deep features (g) appear to be simpler (parsimony) than the original one (a) because of its approaching to the expectation. {This is even more profound in the sense that the compactness did not conflict with the structure preservation, which can be viewed that STM obtains a more similar structure as that of the original manifold than the center loss \cite{centerloss}.} A-Softmax loss \cite{A-Softmax-loss} feature in sub-figure (d) and cosine loss \cite{cosine-loss} feature in sub-figure (f) have a similar distribution like the baseline CNN feature in sub-figure (a). In addition, the center loss in sub-figure (c) or ring loss \cite{ring-loss} in sub-figure (e) appears more scattered, meaning our structure preservation is a better strategy to achieve a good representation. }

%
%
\begin{table}[h]
	\caption{\textcolor{black}{Comparisons with CNNs on the MNIST dataset.}}
	\label{mnist}
	\centering
	\begin{tabular}{|L{3.7cm}|C{3.7cm}|}
		\hline
		\textbf{Models} & \textbf{Results (error rate (\%))}\\
		\hline
		\hline
		Base-CNN (LeNet++ \cite{centerloss}) & 0.73\\
		STN(affine) \cite{stn} & 0.61\\
		Center loss \cite{centerloss} & 0.61\\
		\textcolor{black}{A-Softmax loss \cite{A-Softmax-loss}} & \textcolor{black}{0.49}\\
		\textcolor{black}{Ring loss \cite{ring-loss}} & \textcolor{black}{0.46}\\
		\textcolor{black}{Cosine loss \cite{cosine-loss}} & \textcolor{black}{0.45}\\
		\hline
		STM with LLE (ours) & 0.36\\
		STM with Laplacian (ours) & 0.36\\
		\hline	
	\end{tabular}
\end{table}

In \textcolor{black}{TABLE~\ref{mnist}}, we report the error rates obtained by six different approaches. \textcolor{black}{It can be seen that ours is far lower than the existing approaches including center loss, A-Softmax loss, ring loss and cosine loss, indicating that the manifold loss term indeed increases the discriminative power of the deeply learned features.} In addition, it seems that \textcolor{black}{STM with LLE} performs slightly better depending on parameter selections than \textcolor{black}{STM with Laplacian}. Moreover, the weight calculation for Laplacian is much easier than that of LLE so that we change the notation of \textcolor{black}{STM with Laplacian} to STM and use it in the following experiments. 



\subsection{Natural object recognition}
\textbf{CIFAR \cite{cifa}} dataset is a famous natural image classification benchmark which consists of 60000 32x32 color images in 10 or 100 classes, with 6000 images per class. There are 50000 training images and 10000 test images. 
We follow the same protocol as that of \cite{cifar_rsn}. \textcolor{black}{Seven CNNs including \textcolor{black}{VGG} \cite{vgg}, ResNet \cite{rsn}, WideResNet \cite{cifar_rsn} (or baseline CNN),  center loss \cite{centerloss}, A-Softmax loss \cite{A-Softmax-loss}, ring loss \cite{ring-loss}, and cosine loss \cite{cosine-loss}} are used as baselines on these datasets.

\begin{figure}[h]
	\centering
	\begin{minipage}{0.5\textwidth}
		\centering
		\subfigure[\textcolor{black}{WideResNet (baseline).}]{
			\centering
			\includegraphics[width=0.8\linewidth]{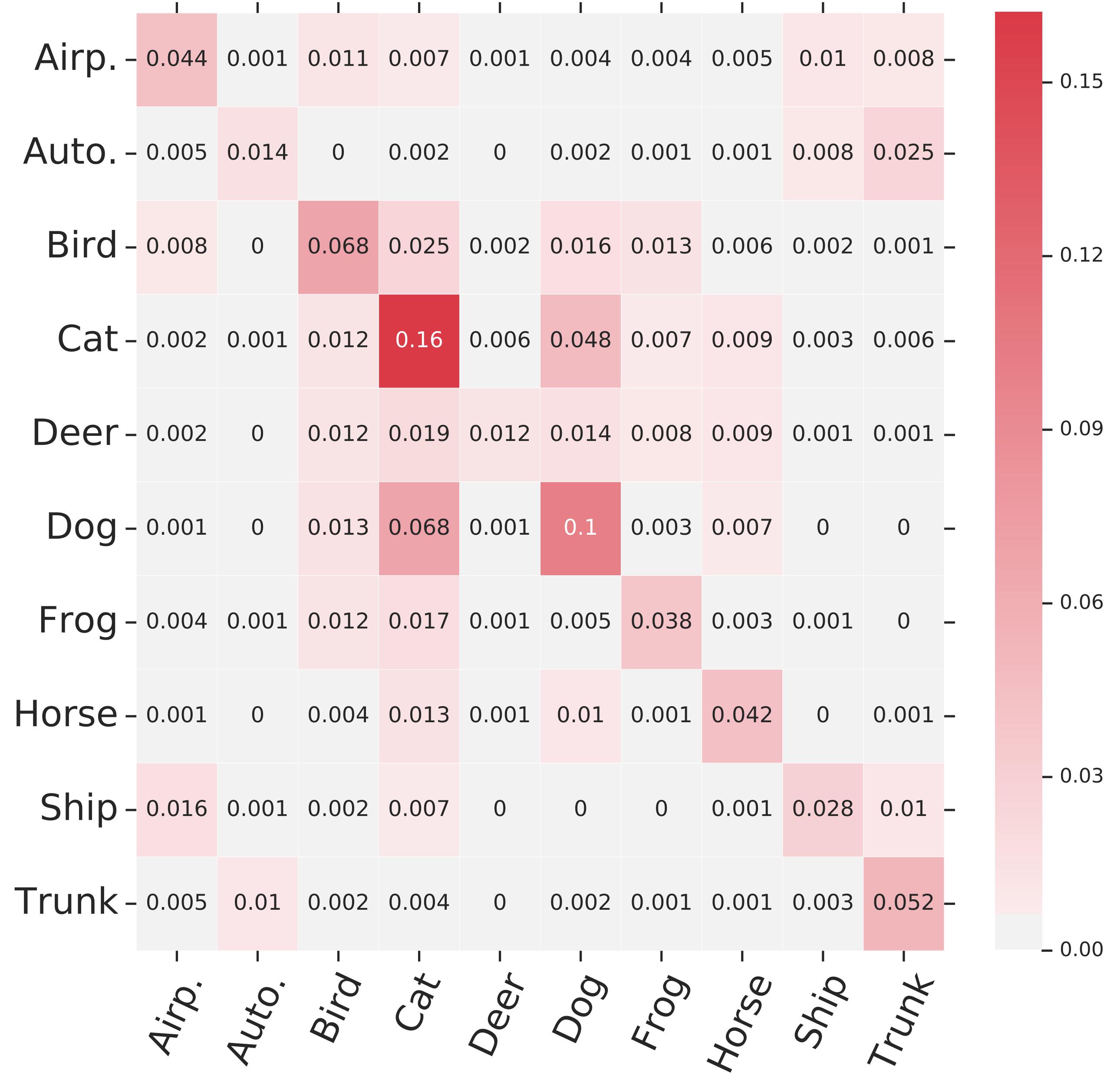}
		}
	\end{minipage}
	\vskip 0.25cm
	\centering
	\begin{minipage}{0.5\textwidth}
		\centering
		\subfigure[\textcolor{black}{STM (ours).}]{
			\centering
			\includegraphics[width=0.8\linewidth]{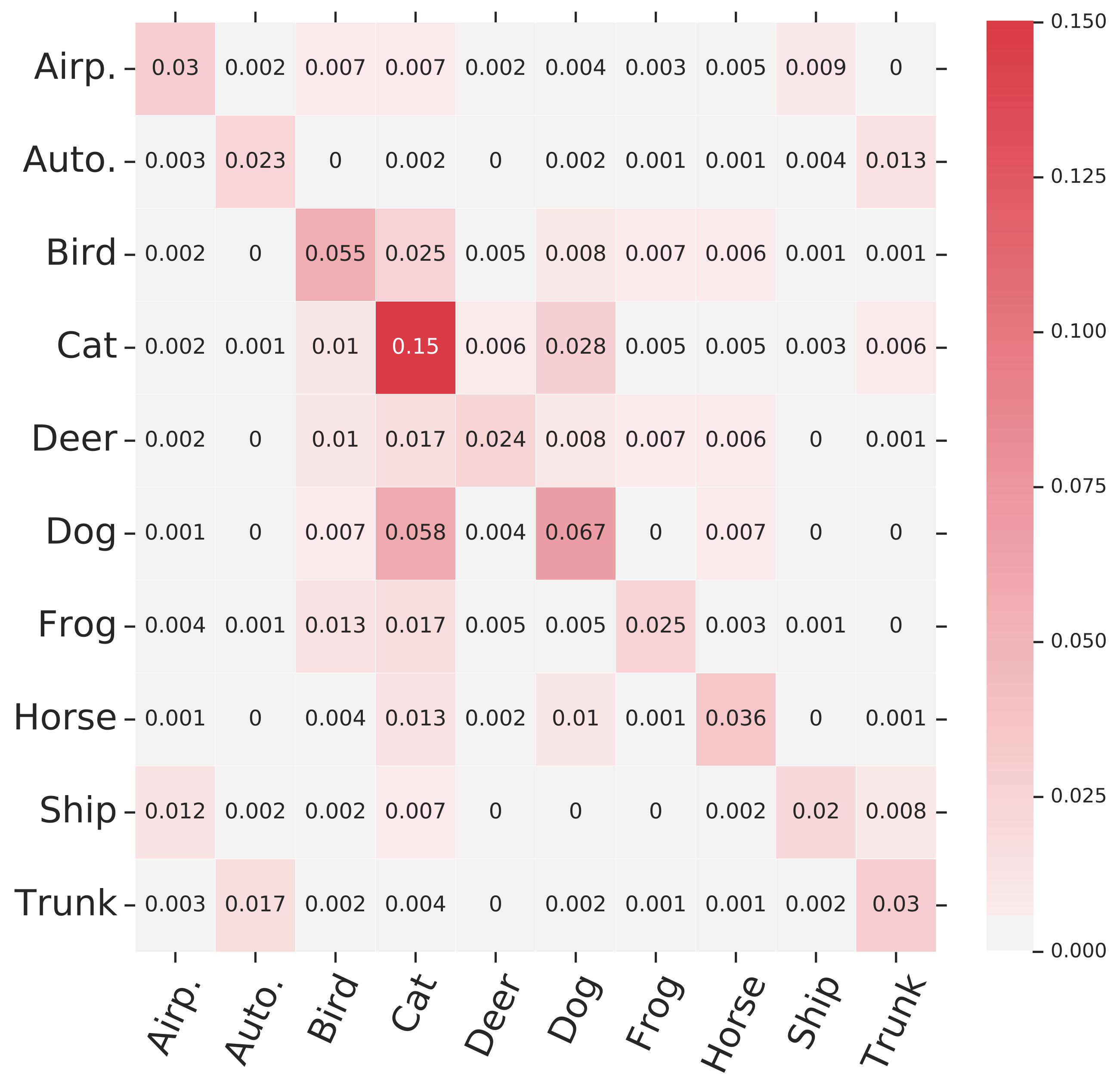}
		}
	\end{minipage}
	\caption{\textcolor{black}{Error distribution analyzing on the CIFAR10 dataset.}}\label{fig:matrix}
\end{figure}

We use a weight decay \textcolor{black}{($\lambda$)} of 0.0001 and momentum of 0.9. These models are trained on two GPUs (Titan XP) with a \textcolor{black}{mini-batch size ($N_b$)} of 128. The learning rate \textcolor{black}{($\mu$)} is started from 0.1, and divided by 10 at 32k and 48k iterations, and the training procedure is terminated at 64k iterations, which is determined on a 45k/5k train/val split. We follow the same data augmentation in \cite{rsn} for training: horizontal flipping is adopted, and a $32 \times 32$ crop is sampled randomly from the image padded by 4 pixels on each side. For testing, we only evaluate the single view of the original $32 \times 32$ image.

\begin{table}[h]
	\caption{\textcolor{black}{Comparisons with CNNs on the CIFAR dataset.}}
	\label{cifar}
	\centering
	\begin{tabular}{|L{3.0cm}|C{2.0cm}|C{2.0cm}|}
		\hline
		\multirow{2}{*}{\textbf{Models}} & \multicolumn{2}{c|}{\textbf{Results on (error rate (\%))}}\\
		\cline{2-3} & \textbf{CIFAR-10} & \textbf{CIFAR-100}\\
		\hline
		\hline
		VGG \cite{vgg} & 6.32 & 28.49\\
		ResNet \cite{rsn} & 6.43 & 25.16\\
		WideResNet \cite{cifar_rsn} & 5.61 & 22.07\\
		Center loss \cite{centerloss} & 5.58 & 22.08\\
		\textcolor{black}{A-Softmax loss \cite{A-Softmax-loss}} & \textcolor{black}{5.56} & \textcolor{black}{22.07}\\
		\textcolor{black}{Ring loss \cite{ring-loss}} & \textcolor{black}{5.54} & \textcolor{black}{22.01}\\
		\textcolor{black}{Cosine loss \cite{cosine-loss}} & \textcolor{black}{5.30} & \textcolor{black}{21.62}\\
		\hline		
		STM (ours) & 4.60 & 20.2\\
		\hline
		
	\end{tabular}
\end{table}
%
%

Our algorithm is also compared with the state-of-the-art algorithms when carrying out the task of image classification. To be fair, the settings for all the algorithms follow WideResNet \cite{cifar_rsn}, which was implemented by us. The results in \textcolor{black}{TABLE~\ref{cifar}} again show that STM significantly improves the baselines (e.g., WideResNet) on both CIFAR10 and CIFAR100 datasets. In \textcolor{black}{Fig.~\ref{fig:matrix}}, we notice that the top-2 classes of being improved in CIFAR10 are dog (34\% higher than baseline WideResNet\footnote{Our implementation in Tensorflow.}), and horse (14\%), in which significant image variations take place. This implies considering the manifold structure in feature learning enhances the capability of handling image variations. In addition, the center loss method \textcolor{black}{(or other three loss methods)} performs worse than STM due to severe variations in the CIFAR datasets.

\subsection{Large size image classification}
The previous experiments are conducted on datasets with small size images. To further show the effectiveness of the proposed STM method, we evaluate it on the ImageNet \cite{Deng2009ImageNet} dataset. Different from MNIST  and CIFAR, ImageNet consists of images with a much higher resolution. In addition, the images usually contain more than one attribute per image, which may have a large impact on the classification accuracy. \textcolor{black}{In this experiment, we firstly choose a 100-class ImageNet 2012 \cite{Deng2009ImageNet} subset for reducing the time for the training a deep model. The 100 classes are selected from the full ImageNet dataset at a step of 10. Similar subset is also applied in \cite{yao2015tiny}. In order to make a more general validation of effectiveness of our STM on large-sized images, we also take the full ImageNet dataset for another test.} 


\begin{table}[h]
	\caption{\textcolor{black}{Comparisons with CNNs on the ImageNet dataset.}}
	\label{imagenet-results}
	\centering
	\begin{tabular}{|L{3.0cm}|C{0.80cm}|C{0.80cm}|C{0.80cm}|C{0.80cm}|}
		\hline
		\multirow{3}{*}{\textbf{Models}} & \multicolumn{4}{c|}{\textbf{Results on (error rate (\%))}}\\
		\cline{2-5} &  \multicolumn{2}{c|}{\textbf{ImageNet-100}} & \multicolumn{2}{c|}{\textcolor{black}{\textbf{ImageNet-Full}}}\\
		\cline{2-5} & Top-1 & Top-5 & Top-1 & Top-5\\
		\hline
		\hline
		ResNet (ResNet-101 \cite{rsn}) & 11.94 & 3.16 & \textcolor{black}{22.44} & \textcolor{black}{6.21} \\
		\textcolor{black}{Center loss \cite{centerloss}} & \textcolor{black}{11.92} & \textcolor{black}{3.15} &  \textcolor{black}{22.31} &  \textcolor{black}{6.18}\\
		\textcolor{black}{A-Softmax loss \cite{A-Softmax-loss}} & \textcolor{black}{11.28} & \textcolor{black}{3.12} & \textcolor{black}{22.15} & \textcolor{black}{6.16}\\
		\textcolor{black}{Ring loss \cite{ring-loss}} & \textcolor{black}{11.13} & \textcolor{black}{3.10} & \textcolor{black}{22.10} &  \textcolor{black}{6.09}\\
		\textcolor{black}{Cosine loss \cite{cosine-loss}} & \textcolor{black}{11.28} & \textcolor{black}{3.09} & \textcolor{black}{22.14} &\textcolor{black}{6.15} \\
		\hline		
		STM (ours) & 10.67 & 2.94 & \textcolor{black}{21.56} & \textcolor{black}{5.77} \\
		\hline
	\end{tabular}
\end{table}

For the ImageNet-100 and ImageNet-Full experiment, we use the same model as the baseline ResNet (ResNet-101 \cite{rsn}) model, and the setting is the same as the previous experiments. Both methods are trained after 120 epochs. The learning rate \textcolor{black}{($\mu$)} is initialized as 0.1 and decreases to 1/10 times per 30 epochs. Top-1 and Top-5 errors are used as evaluation metrics. The test errors are shown in \textcolor{black}{TABLE~\ref{imagenet-results}}. \textcolor{black}{Compared to the ResNet baseline, our STM achieves a better classification performances (i.e., Top-5 error: 2.94\% vs. 3.16\% for ImageNet-100, 5.77\% vs. 6.21\% for ImageNet-Full, Top-1 error: 10.67\% vs. 11.94\% for ImageNet-100, 21.56\% vs. 22.44\% for ImageNet-Full) with almost the same parameters (44.54M). With respect to other baselines, such as center loss, A-Softmax loss, ring loss, and cosine loss, STM also achieves remarkable improvements.} Considering the large variations in ImageNet, STM can still achieve a better performance than ResNet, and we believe that manifold loss is really effective. 


%
















\subsection{Object tracking}
In this section, we evaluate the performance of STM on the tracking problem based on 50 sequences from the commonly used tracking \textcolor{black}{\textit{object tracking benchmark} (OTB) dataset} \cite{otb}. 

\begin{figure}[h]
	\centering
	\begin{minipage}{0.5\textwidth}
		\centering
		\subfigure[Precision plots.]{
			\centering
			\includegraphics[width=0.80\linewidth]{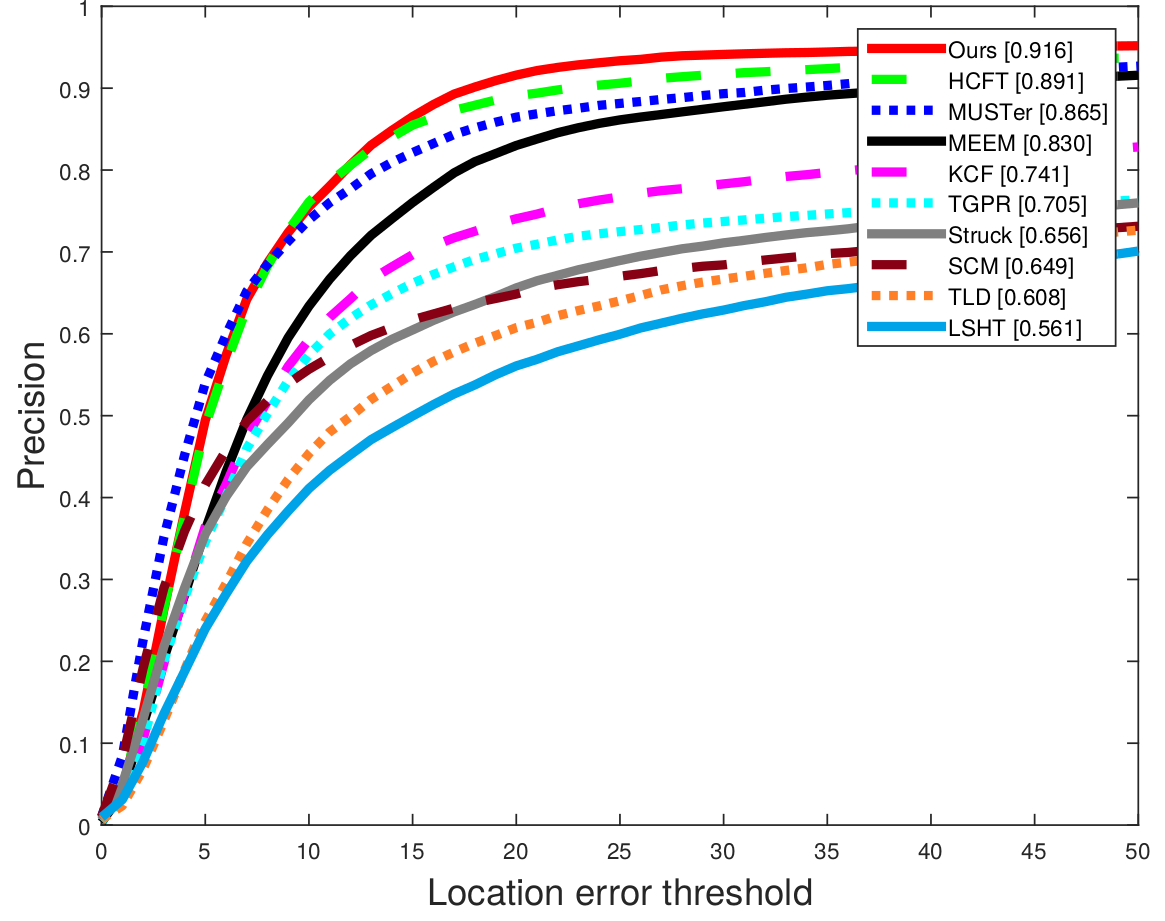}
		}
	\end{minipage}
	\centering
	\begin{minipage}{0.5\textwidth}
		\centering
		\subfigure[Success plots.]{
			\centering
			\includegraphics[width=0.80\linewidth]{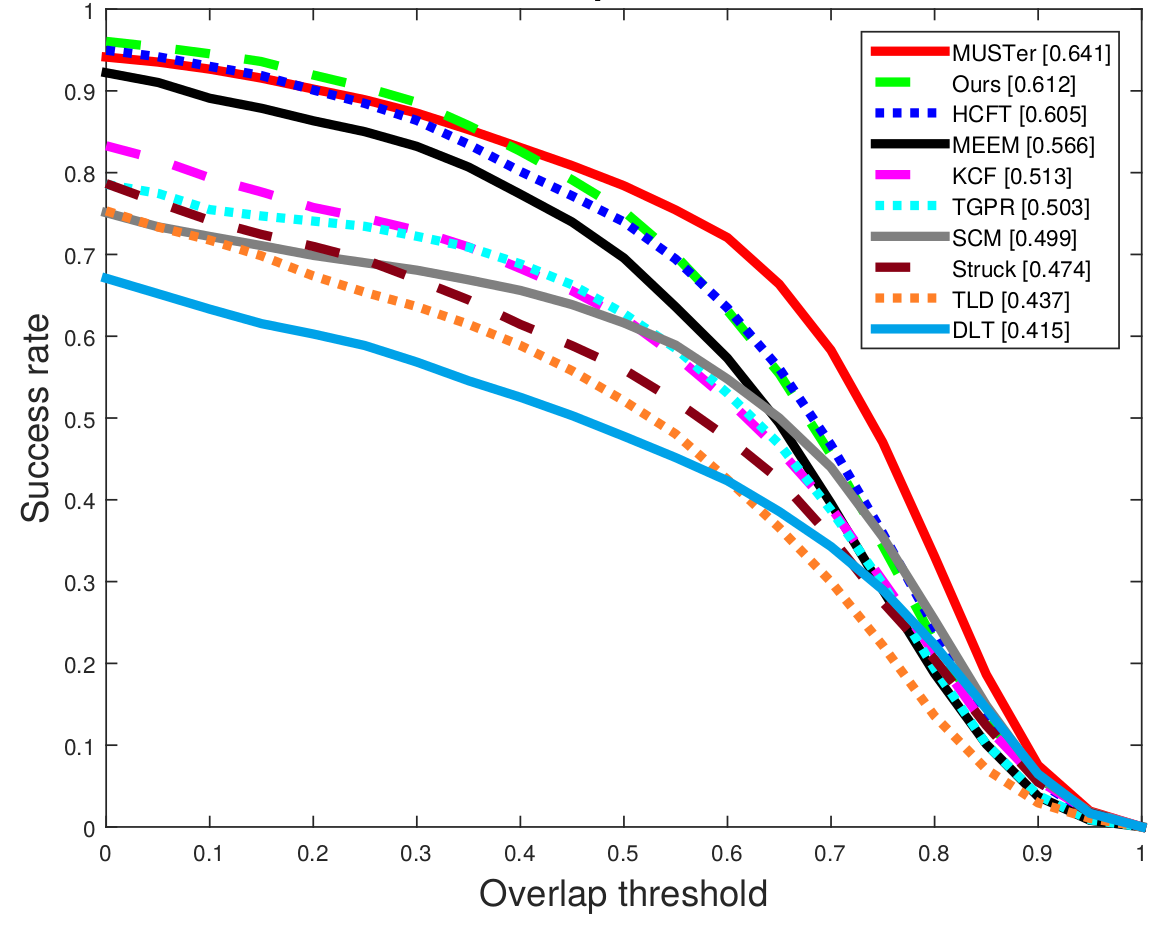}
		}
	\end{minipage}
	\caption{\textcolor{black}{Precision and success plots on the OTB dataset.}}\label{fig:overall}
\end{figure}

\textbf{OTB} \cite{otb} is a large dataset with ground-truth object positions and extents for tracking and introduces the sequence attributes for the performance analysis. They integrate most of the publicly available trackers into one code library with the uniform input and output formats to facilitate large-scale performance evaluation. The performances of most tracking algorithms are included on 50 sequences with different initialization settings. In this tracking benchmark \cite{otb}, each sequence is manually tagged with different attributes, such as \emph{illumination variations}, \emph{scale variations}, \emph{occlusions}, \emph{deformations}, \emph{motion blur}, \emph{abrupt motion}, \emph{in-plane rotation}, \emph{out-of-plane rotation}, \emph{out-of-view}, \emph{background clutters} and \emph{low resolution}, indicating what kind of challenges exist in the video. 

We implement our STM model based on \textcolor{black}{VGG-19 \cite{vgg}} with two outputs for each sequence separately. We randomly collect 50 positive and 200 negative samples for each frame from VOT13, VOT14, and VOT15\footnote{http://www.votchallenge.net/}, where the positive and negative examples have $0.7$ and $0.5$ IoU overlap ratios with ground-truth bounding boxes, respectively. Noted that we remove the overlapped sequences with OTB from the trained databases. In the tracking, we used the same strategy as that of \cite{C.Ma,C.Ma1}, which learns a discriminative classifier and estimates the translation of target objects by searching for the maximum value of correlation response map. Similar to KCF \cite{kcf}, using the set of correlation response maps based on deep features can hierarchically infer the target translation at each layer, i.e., the location of the maximum value in the last layer is used as a regularization to search for the maximum value of the earlier layer. Our STM tracker can be generated by simply replacing the deep model of \cite{C.Ma,C.Ma1}. Regarding the comparison, our baseline algorithms mainly consist of correlation filters or deep learning based trackers, such as KCF, FCNT, and Cf+CNN \cite{C.Ma1}.

\textcolor{black}{In Fig.~\ref{fig:overall}}, we plot the precision against location error curve, which measures the ratio of successful tracking frames when the threshold of allowed location errors is changed. Here, the location error (x-axis, in pixel) on the plot implies the distance between the bounding box center and the ground-truth. For ease of comparison, we also include the plots of several baseline trackers in the figure. 

%
\begin{table}[htbp!]
	\caption{\textcolor{black}{Comparisons with state-of-the-art trackers on the OTB dataset.}}
	\label{otbtable}
	\centering
	\begin{tabular}{|L{2.75cm}|C{2.15cm}|C{2.15cm}|}
		\hline
		\multirow{2}{*}{\textbf{Models}} & \multicolumn{2}{c|}{\textbf{Results}}\\
		\cline{2-3} & \textbf{Precision (\%)} & \textbf{Success rate (\%)}\\
		\hline
		\hline
		FCNT \cite{L.Wang}  & 85.7 & 47.2\\
		KCF \cite{kcf}      & 74.1 & 51.3\\
		Cf+CNN \cite{C.Ma1} & 90.7 & 61.1\\
		HCFT \cite{C.Ma}    & 89.1 & 60.5\\
		MEEM \cite{meem}    & 83.0 & 56.6\\
		DSST \cite{DSST}    & 73.9 & 50.5\\		
		\hline		
		STM (ours) & 91.6 & 61.2\\
		\hline
		
	\end{tabular}
\end{table}

As can be seen in \textcolor{black}{TABLE~\ref{otbtable}}, the STM and KCF achieve 61.2\% and 51.3\% based on the average success rate, while HCFT and MEEM trackers respectively achieve 60.5\% and 56.6\%. In terms of precision, STM and KCF respectively achieve 91.6\% and 74.1\% when the threshold is set to 20. Moreover, the STM and baseline HCFT obtain 91.6\% and 89.1\% respectively, which further confirms that the proposed deep model is effective on object tracking. We also compare with cf+CNN, one of the latest variants of KCF, and the results show that STM still achieves performance improvement in terms of precision. It is believed that the special strategy used in cf+CNN can also be used to further improve STM. All the above observations clearly demonstrate that imposing the manifold prior constraint during the feature learning helps generate more robust features for tracking, thus enabling its superiority over the state-of-the-art trackers. 
	


\begin{figure}[h]
	\centering
	\begin{minipage}{0.5\textwidth}
		\centering
		\subfigure[Precision plots for scale variation.]{
			\centering
			\includegraphics[width=0.80\linewidth]{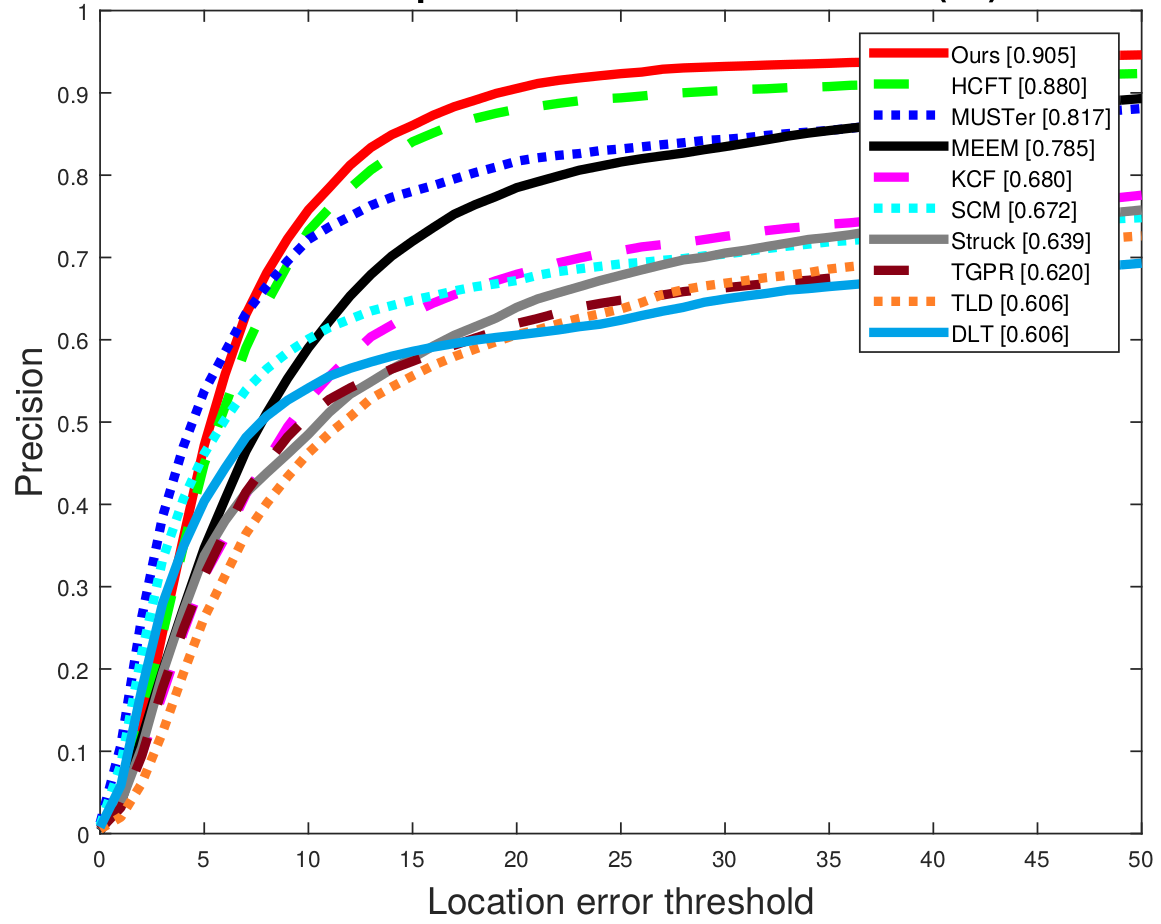}
		}
	\end{minipage}
	\centering
	\begin{minipage}{0.5\textwidth}
		\centering
		\subfigure[Precision plots for illumination variation.]{
			\centering
			\includegraphics[width=0.80\linewidth]{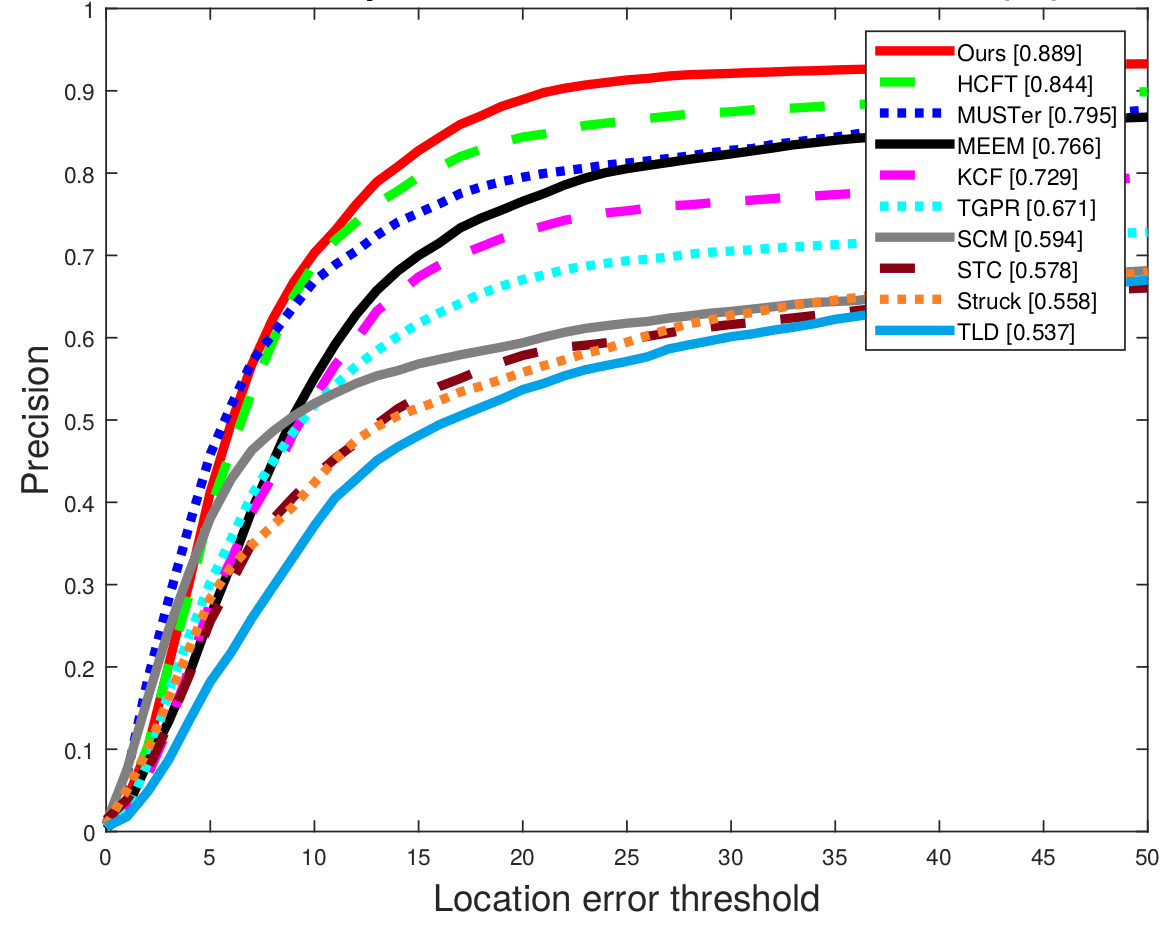}
		}
	\end{minipage}
	\caption{\textcolor{black}{Precision plots for two attributes (scale variation and illumination variation) on the OTB dataset.}}
	\label{attributes}
\end{figure}

Here, we also show the scale variation and lighting attributes in \textcolor{black}{Fig.~\ref{attributes}}, where STM performs much better than other trackers again. \textcolor{black}{The reason for this phenomenon is that the manifold structure of the learned features is data-dependent. This property leads the learned features can handle nonlinear of variations in object tracking.} Again, STM shows its super capability of handling severe variations.  \textcolor{black}{The experimental results for full set of plots generated by the benchmark toolbox are reported in Fig.~\ref{precision-attributes-all} and Fig.~\ref{success-attributes-all} in Appendix \ref{app-OTB}.}

\section{Conclusion}\label{Sec.5}
In this paper, we have presented a new concept for representation learning that data structure preservation can help feature learning process converge at the representation expectation. Thus, we open up a possible way to learn deep features which are robust to the variations of the input data because of theoretical convergence. The proposed STM method formulates the data structure preserving into an objective function optimization problem with constraint, which can be solved via the BP algorithm. Extensive experiments and comparisons on the commonly used benchmarks show that the proposed method significantly improved the CNN performance, and achieved a better performance than the state-of-the-arts.

\bibliography{References-TIP}
\bibliographystyle{ieee}

\appendices
\section{Algorithms}\label{app-algorithms}

\subsection{\textcolor{black}{Algorithm of refreshing data buffer and feature buffer}}\label{app-algorithm-refresh}
\textcolor{black}{\textbf{Algorithm~\ref{algo-refresh}} shows the algorithm of refreshing data buffer and feature buffer. The input of this algorithm includes the old data buffer $D'_{B}$, the old feature buffer $\hat{F}'$, the candidate data buffer $D_{N_b}$, the candidate feature buffer $F$, the manifold loss balance parameter $\gamma$, the number of neighbors $k_l$, the manifold loss balance parameter $\gamma$, the indicator ${m}\in\{{lle}, {lap}\}$ for denoting which manifold method is used in training. The output of this algorithm is the refreshed data buffer $D_{B}$ and the refreshed feature buffer $\hat{F}$.} 

\begin{algorithm}[h]
	\SetKwInput{Input}{Input}
	\SetKwInput{Output}{Output}
	\SetKw{KwGoTo}{go to line}
	
	\Input{\textcolor{black}{$D'_{B}$, $\hat{F}_{B}'$, $D_{N_b}$, $F_{N_b}$, $\gamma$, $k_l$, $m$;}}
	\Output{\textcolor{black}{$D_{B}$, $\hat{F}$;}}
	$\mathcal{J}' \gets m(D_{B}', \hat{F}', \gamma, k_l)$\;
	$\mathcal{J} \gets m(D_{N_b}, {F}, \gamma, k_l)$\;	
	
	\eIf{$\mathcal{J} \ge \mathcal{J}'$}{
		\textcolor{black}{$\{D_{B}, \hat{F}_{B}\} \gets \{D_{B}', \hat{F}'_{B}\}$\;
		\KwGoTo \ref{marker}\;}
	}
	{
		\eIf{$B \le N_b$}{
			\textcolor{black}{$\{D_{B}, \hat{F}_{B}\} \sim \mathcal{G}(D_{N_b}', \hat{F}'_{N_b})$\;}
		}
		{
			\textcolor{black}{$D_{B} \gets D'_{B-N_b} \bigcup D_{N_b}$\;}
			\textcolor{black}{$\hat{F}_{B} \gets \hat{F}'_{B-N_b} \bigcup \hat{F}_{N_b}$\;}
		}
	}

	\textcolor{black}{\Return $\{D_{B}, \hat{F}_{B}\}$\;} \label{marker}
	
	\caption{\textcolor{black}{Refreshing data and feature buffer, denoted as $\{D_{B}, \hat{F}_{B}\} = \text{\textit{refresh}}(D'_{B}, \hat{F}'_{B}, D_{N_b}, F_{N_b}, \gamma, $ $k_l, m)$.}}\label{algo-refresh} 
\end{algorithm}

\textcolor{black}{The procedures for refreshing the buffers include following three main steps. }
	
\textcolor{black}{\textbf{Step 1. Calculation of the condition for updated buffers, line 1-2.} Line 1 is used to calculate the manifold loss for the old data buffer $D'_{B}$ and the old feature buffer $\hat{F}'$, and line 2 for the candidate buffers. The complexities of this step depends on the function $lle(\cdot)$ (as shown in \textbf{Algorithm~\ref{algo-lle}}) or the function $lap(\cdot)$  (as shown in \textbf{Algorithm~\ref{algo-lap}}). We denote the complexities of this step by $\mathcal{T}_m$ and $\mathcal{S}_m$.}

\textcolor{black}{\textbf{Step 2. The buffers updating, line 3-13.} If the manifold loss of the candidate data buffer and the candidate feature buffer is larger than or equal to the manifold loss of the old data buffer and the old feature buffer (line 3), then not change the buffers (line 4) and go to line 14, i.e., \textbf{Step 3}; otherwise, refresh the buffers (line 6-13). If the buffer size $B$ (i.e., $B=k_0\times|Y|$) is lesser than or equal to the mini-batch size $N_b$ (line 12), then randomly choose $B$ samples from candidate buffers (line 13); otherwise, remove $N_b$ samples from the old buffer and set the new buffer with all samples from candidate buffer and the rest samples from the old buffer (line 10, 11). The time complexity and space complexity of this step are both equal to the buffer complexity, i.e., $O(B\times(|D|+|F|))$.}

\textcolor{black}{\textbf{Step 3. Return the updated buffers, line 14.} Return the refreshed buffers $\{D_{B}, \hat{F}_{B}\}$.}

\textcolor{black}{In summary, the time complexity of this algorithm is}
\textcolor{black}{\begin{align}\label{eq-time-comlexity-refresh}
	\mathcal{T}_{r} = O(B\times(|D|+ |F|)) + \mathcal{T}_{m},
	\end{align}}
\textcolor{black}{and the space complexity in each mini-batch training is}
\textcolor{black}{\begin{align}\label{eq-space-comlexity-refresh}
	\mathcal{S}_{r} = O(B\times(|D|+ |F|)) + \mathcal{S}_{m}.
	\end{align}}

\subsection{\textcolor{black}{Algorithm of calculating the LLE manifold loss}}\label{app-algorithm-lle}
\textcolor{black}{\textbf{Algorithm~\ref{algo-lle}} shows the algorithm of calculating the LLE manifold loss. The input of this algorithm includes data buffer $D_B$, feature buffer $\hat{F}$, the manifold loss balance parameter $\gamma$, and the number of neighbors $k_l$. The output of this algorithm is the LLE manifold loss $\mathcal{J}$.}

\begin{algorithm}[h]
	\SetKwInput{Input}{Input}
	\SetKwInput{Output}{Output}
	
	\Input{\textcolor{black}{$D_{B}$, $\hat{F}$, $\gamma$, $k_l$;}}
	\Output{\textcolor{black}{$\mathcal{J}$;}}
	
	\textcolor{black}{$N \gets [N_{i} = k\text{-NN}(X_i, k_l)|X_i\in D_{B}]$, where $N_i=[n_{1,i}, n_{2,i}, ..., n_{k_l,i}]$\;}
	
	\textcolor{black}{\eIf{$k_l > |X_i|$}{$\epsilon \gets 10^{-4}$\;}{$\epsilon \gets 0$\;}}
	\textcolor{black}{$A^{{\cal M}} \gets [\alpha_{i,j}=0|i,j=1,2, ..., B]$\;}
	\textcolor{black}{$W \gets [w_{k,i}=0|k=1,2,...,k_l; i=1,2, ..., B]$\;}
	
	%
	%
	
	\For{$i = 1, 2, ..., B$}{
		\textcolor{black}{$z = \{X_k-X_i|k=n_{1,i}, n_{2,i}, ..., n_{k_l,i}\}$\;}
		\textcolor{black}{$Z \gets z^{\top}z$\;}
		
		\textcolor{black}{$Z \gets Z+\epsilon\times \text{trace}(Z)\times \text{eye}(k_l, k_l)$\;}
		
		\textcolor{black}{$W[:,i] \gets Z\backslash\text{ones}(k_l, 1)$\;}
		\For{$j=1, 2, ..., k_l$}{
			\textcolor{black}{$w_{j,i} \gets w_{j,i}/\sum_{j=1}^{k_l}w_{j,i}$\;}
			\textcolor{black}{$\alpha_{n_{j,i}, i} \gets w_{j,i}$\;}
		}
	}
	
	\textcolor{black}{$\mathcal{J} \gets \frac{\gamma}{2|\hat{F}|} \sum_{\hat{F}_i \in \hat{F}} \| {\hat{F}_i - \hat{F}  A^{\cal M}_{i}}\|^2$\;}	
	
	\textcolor{black}{\Return  $\mathcal{J}$\;}
	
	\caption{\textcolor{black}{Calculating the LLE manifold loss, denoted as $\mathcal{J} = lle(D_{B}, \hat{F}, \gamma, k_l)$.}}\label{algo-lle}
\end{algorithm}

\textcolor{black}{The procedures for calculating the LLE manifold loss include following four main steps.}

\textcolor{black}{\textbf{Step 1. Setting the neighbors' id, line 1.} Set the neighbors' id by k-nearest neightbor algorithm $k\text{-NN}(\cdot)$. The time complexity and space complexity of this step are both equal to $O(B^2\times|D|)$.}

\textcolor{black}{\textbf{Step 2. Calculating the weight for each neighbor, line 2-18.} Calculate the weight $\alpha_{i,j}$ for each neighbor pair $X_i$, $X_j$. The time complexity and space complexity of this step are both equal to $O(B^2\times|D|)$.}

\textcolor{black}{\textbf{Step 3. Calculating the manifold loss, line 19.} Calculating the manifold loss $\mathcal{J}$ based on the feature buffer $\hat{F}$ and the weight matrix $A^{\mathcal{M}}$. The time complexity of this step is $O(B^2\times|F|)$, and the space complexity of this step is $O(1)$.}

\textcolor{black}{\textbf{Step 4. Return the manifold loss, line 20.} Return the manifold loss $\mathcal{J}$.}

\textcolor{black}{In summary, the time complexity of this algorithm is}
\textcolor{black}{\begin{align}\label{eq-time-comlexity-lle}
	\mathcal{T}_{lle} = O(B^{2}\times(|D|+|F|)),
	\end{align}}
\textcolor{black}{and the space complexity in each mini-batch training is}
\textcolor{black}{\begin{align}\label{eq-space-comlexity-lle}
	\mathcal{S}_{lle} =  O(B^{2}\times|D|).
	\end{align}}

\subsection{\textcolor{black}{Algorithm of calculating the Laplacian manifold loss}}\label{app-algorithm-lap}
\textcolor{black}{\textbf{Algorithm~\ref{algo-lap}} shows the algorithm of calculating the Laplacian manifold loss. The input of this algorithm includes data buffer $D_B$, feature buffer $\hat{F}$, the manifold loss balance parameter $\gamma$, and the number of neighbors $k_l$. The output of this algorithm is the Laplacian manifold loss $\mathcal{J}$.}

\begin{algorithm}[h]
	\SetKwInput{Input}{Input}
	\SetKwInput{Output}{Output}
	
	\Input{\textcolor{black}{$D_{B}$, $\hat{F}$, $\gamma$, $k_l$;}}
	\Output{\textcolor{black}{$\mathcal{J}$;}}
	
	\textcolor{black}{$N \gets [N_{i} = k\text{-NN}(X_i, k_l)|X_i\in D_{B}]$, where $N_i=[n_{1,i}, n_{2,i}, ..., n_{k_l,i}]$\;}
			
	\textcolor{black}{$B^{{\cal M}} \gets [\beta_{i,j}=0|i,j=1,2, ..., B]$\;}
	
	\textcolor{black}{$\epsilon \gets \max_{X_i, X_j \in D_B}\|X_{i}-X_{j}\|^{2}/B^2$\;}
	
	\For{$X_i \in D_B$}{
		\For{$j\in N_i$}{
			\textcolor{black}{$\beta_{j,i} \gets \exp(\|X_{i}-X_{j}\|^{2}/\epsilon)$\;}
		}
	
		\textcolor{black}{$\beta_i \gets \sum_{j\in N_i}\beta_{j,i}$\;}
		\For{$j\in N_i$}{
			\textcolor{black}{$\beta_{j,i} \gets \beta_{j,i}/\beta_i$\;}
		}

	}
		
	
	\textcolor{black}{$\mathcal{J} \gets	 \frac{\gamma}{2|\hat{F}|^2} \sum_{\hat{F}_i, \hat{F}_j \in \hat{F}} \|\hat{F}_i^{} - \hat{F}_j^{}\|^2 B^{\cal M}_{ij}$\;}
	
	\textcolor{black}{\Return $\mathcal{J}$\;}
	
	\caption{\textcolor{black}{Calculating the Laplacian manifold loss, denoted as $\mathcal{J} = lap(D_B, \hat{F}, \gamma, k_l)$.}}\label{algo-lap}
\end{algorithm}

\textcolor{black}{The procedures for calculating the Laplacian manifold loss include following three main steps.}
 
\textcolor{black}{\textbf{Step 1. Calculating the weight for each neighbor, line 1-12.} Calculate the weight $\beta_{i,j}$ for each neighbor pair $X_i$, $X_j$.  The time complexity and space complexity of this step are both equal to $O(B^2\times|D|)$.}

\textcolor{black}{\textbf{Step 2. Calculating the manifold loss, line 13.} Calculating the manifold loss $\mathcal{J}$ based on the feature buffer $\hat{F}$ and the weight matrix $B^{\mathcal{M}}$. The time complexity of this step is $O(B^2\times|F|)$, and the space complexity of this step is $O(1)$.}

\textcolor{black}{\textbf{Step 3. Return the manifold loss, line 14.} Return the manifold loss $\mathcal{J}$.}

\textcolor{black}{In summary, the time complexity of this algorithm is}
\textcolor{black}{\begin{align}\label{eq-time-comlexity-lap}
	\mathcal{T}_{lap} = O(B^{2}\times(|D|+|F|)),
	\end{align}}
\textcolor{black}{and the space complexity in each mini-batch training is}
\textcolor{black}{\begin{align}\label{eq-space-comlexity-lap}
	\mathcal{S}_{lap} =  O(B^{2}\times|D|).
	\end{align}}

\section{Proof of Lemma 1}\label{app-proof-lemma-1}

\textbf{Lemma 1:} If $x_{1},x_{2},...,x_{n}$ satisfies Definition 1, then: 
\begin{equation}
\textcolor{black}{
	P(|\bar{x}-E(\bar{x})| \ge \frac{\lambda}{n} ) \le 2 \exp (\frac{-\lambda^2}{2\sum_{i=1}^{n}{c_i^2}}),
}
\end{equation}
where \textcolor{black}{$\bar{x}$} is  the average of $x_{1},x_{2},...,x_{n}$. 

\begin{IEEEproof} For a fixed  \textcolor{black}{$t$} ($t \ge 0$), the function  $e^{t {y} }$ of the variable ${y}$ is convex in the interval  $[-g,g]$ with $g \ge 0$.We draw a line between the two endpoints points $(-g,e^{-t g})$ and $(g,e^{t g})$. The curve of   $e^{t {y}}$ lies entirely below this line. Thus,
	
	\begin{equation}
	\label{3}
	e^{t {y}} \le \frac{g- {y}}{2g} e^{-t g} + \frac{g+ {y}}{2g} e^{tg}\textcolor{black}{.}
	\end{equation}
	
	According to \textcolor{black}{Eq.~(\ref{3})} and $\|x_{i-1}-x_{i}\| \le c_{i} $ (actually $|x_{i-1}-x_{i}| \le c_{i} $), we have: 
	\begin{equation}
	\label{4}
	\textcolor{black}{
		\begin{aligned}
		E(e^{t(x_{i}-x_{i-1})}|x_{[1:i-1]}) &\le E(\frac{(e^{tc_{i}}-e^{-tc_{i}})(x_{i}-x_{i-1})}{2c_{i}}|x_{[1:i-1]}) +\\ &E(\frac{e^{tc_{i}}+e^{-tc_{i}}}{2} |x_{[1:i-1]})\\
		&= \frac{e^{tc_{i}}+e^{-tc_{i}}}{2},
		\end{aligned}}
	\end{equation}
	where \textcolor{black}{$x_{[1:i-1]}=\{x_1,x_2,...,x_{i-1}\}$}. Based on \textcolor{black}{\textbf{Definition~\ref{definition-1}}}, we have:
	\textcolor{black}{$$E(\frac{(e^{tc_{i}}-e^{-tc_{i}})(x_{i}-x_{i-1})}{2c_{i}}|x_{[1:i-1]}) = 0.$$}
	Using the Taylor expansion, we  have:	
	\begin{equation}
	\frac{e^{tc_{i}}+e^{-tc_{i}}}{2} \le \textcolor{black}{\exp(\frac{t^{2} c^{2}_{i}}{2})}.
	\end{equation}
	With the condition  $E(e^{tx_{i-1}}|\textcolor{black}{x_{[1:i-1]}})=e^{tx_{i-1}} $, we have:
	
	\begin{equation}\label{6}\textcolor{black}{
		E(e^{tx_{i}}|x_{[1:i-1]}) \le \exp(\frac{t^{2} c^{2}_{i}}{2}) e^{tx_{i-1}}.}
	\end{equation}
	Inductively, we have:
	\begin{equation}
	\label{7}
	\textcolor{black}{
		\begin{aligned}
		E(e^{tx})&=E(E(e^{tx_{n}}|x_{[1:n-1]}))\\
		&\le \exp(\frac{t^{2} c^{2}_{n}}{2}) E(e^{tx_{n-1}})\le \cdots \le \prod_{i=1}^{n}{\exp(\frac{t^{2} c^{2}_{i}}{2}) E(e^{tx_{i}})}\\
		&=\exp(\frac{1}{2} t^{2} \sum_{i=1}^{n}{c^{2}_{i}}) \exp(tE(x)),
		\end{aligned}
	}
	\end{equation}
	where ${x}$ is the sum of the input samples. 
	%
	%
	%
	%
	%
	%
	According to  Markov's inequality, we have:
	\begin{equation}
	\textcolor{black}{
		\begin{aligned}
		P(x \ge E(x) + \lambda)&=P(\exp(t(x-E(x))) \ge e^{t \lambda})\\
		&\le e^{-t \lambda} E(e^{t(x-E(x))})\\
		&\le e^{-t \lambda} \exp(\frac{t^{2}\sum_{i=1}^{n}{c^{2}_{i}}}{2})\\
		&=\exp(-t \lambda + \frac{1}{2}t^{2}\sum_{i=1}^{n}{c^{2}_{i}}).
		\end{aligned}
	}
	\end{equation}
	We choose \textcolor{black}{$t= \lambda/\sum_{i=1}^{n}{c^{2}_{i}}$} (in order to minimize the above expression), and have:
	\begin{equation}
	\textcolor{black}{
		\begin{aligned}
		P(x \ge E(x)+ \lambda) &\le \exp(-t \lambda + \frac{1}{2}t^{2}\sum_{i=1}^{n}{c^{2}_{i}})\\
		&=\exp(\frac{- \lambda ^{2}}{2 \sum_{i=1}^{n}{c^{2}_{i}}}).
		\end{aligned}}
	\end{equation}
	To derive a similar lower bound, we consider $-x_{i}$ instead of  $x_{i}$ in the preceding proof. Then we obtain the following bound for the lower tail:
	
	\begin{equation}\textcolor{black}{
		P(x \le E(x)- \lambda) \le \exp(\frac{- \lambda ^{2}}{2 \sum_{i=1}^{n}{c^{2}_{i}}})}.
	\end{equation}
	So, we have:
	
	\begin{equation}
	\textcolor{black}{
		P(|\bar{x}-E(\bar{x})| \ge \frac{\lambda}{n}) \le 2 \exp(\frac{- \lambda ^{2}}{2 \sum_{i=1}^{n}{c^{2}_{i}}}),}
	\end{equation}
	where \textcolor{black}{$\bar{x} = \frac{1}{n}\sum_{i=1}^{n}x_{i}$} is the average.
	Thus, the theorem is proved.	
\end{IEEEproof}


\section{Comparing Precision and Success Plots for All Attributes of the OTB Dataset}\label{app-OTB}
\begin{figure}[h]
	\centering
	\begin{minipage}{0.49\textwidth}
		\centering
		\includegraphics[width=0.95\linewidth]{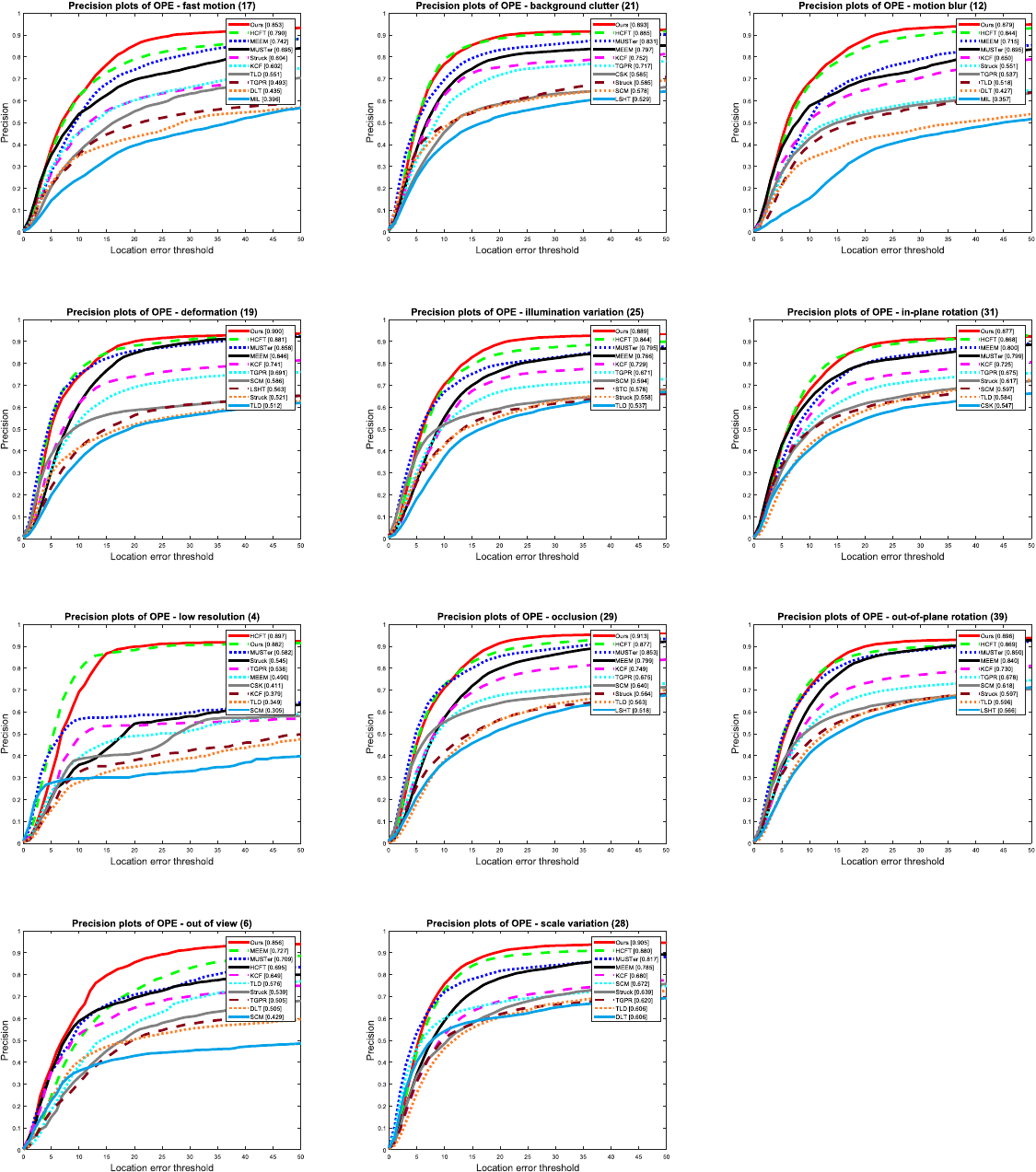}
	\end{minipage}
	\caption{\textcolor{black}{Precision plots for all attributes of the OTB dataset.}}
	\label{precision-attributes-all}
\end{figure}

\begin{figure}[h]
	\centering
	\begin{minipage}{0.49\textwidth}
		\centering
		\includegraphics[width=0.95\linewidth]{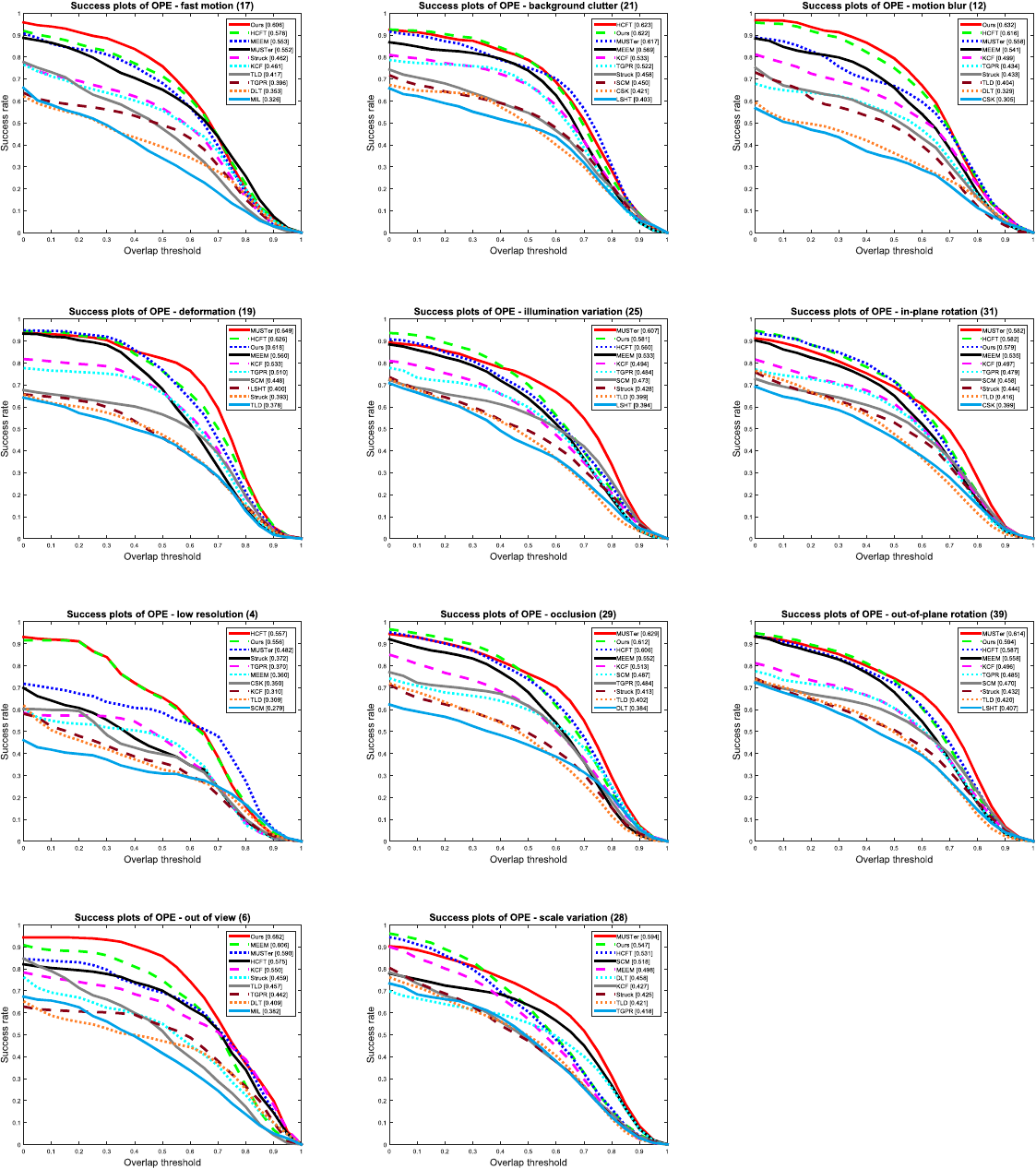}
	\end{minipage}
	\caption{\textcolor{black}{Success plots for all attributes of the OTB dataset.}}
	\label{success-attributes-all}
\end{figure}

\textcolor{black}{The experimental results (precision and success) for full set of plots generated by the benchmark toolbox of OTB \cite{otb} are reported in Fig.~\ref{precision-attributes-all} and Fig.~\ref{success-attributes-all}, where STM performs  better than other trackers.}

%

\end{document}